\pdfoutput=1
\documentclass{article}
\usepackage{iclr2027_conference,times}

\usepackage{amsmath,amsfonts,bm}

\def\eqref#1{equation~\ref{#1}}

\def\1{\bm{1}}

\DeclareMathAlphabet{\mathsfit}{\encodingdefault}{\sfdefault}{m}{sl}
\SetMathAlphabet{\mathsfit}{bold}{\encodingdefault}{\sfdefault}{bx}{n}

\usepackage{hyperref}
\usepackage{url}
\usepackage{graphicx}
\graphicspath{{figures/}}
\usepackage{amsmath,amssymb}
\usepackage{booktabs}
\usepackage{wrapfig}
\usepackage{xcolor}
\usepackage{tcolorbox}
\tcbuselibrary{breakable,skins,listings}

\newcommand{\model}{TimeOmni-o1}
\newcommand{\dataset}{MMTR}

\newcommand{\pct}[1]{{\scriptsize(#1\%)}}
\newcommand{\cmark}{\checkmark}
\newcommand{\xmark}{$\times$}
\newtcblisting[auto counter]{promptbox}[2][]{%
  enhanced, breakable, listing only,
  colback=black!3, colframe=black!55, boxrule=0.5pt, arc=1.5pt,
  left=3pt, right=3pt, top=2pt, bottom=2pt,
  fonttitle=\bfseries\small, coltitle=black, colbacktitle=black!10,
  title={Prompt~\thetcbcounter: #2},
  listing options={basicstyle=\ttfamily\scriptsize, breaklines=true, columns=fullflexible,
    keepspaces=true, literate={—}{{{\rmfamily\textemdash}}}1 {`}{{\textasciigrave}}1},
  #1}
\newtcolorbox{samplebox}[2][]{%
  enhanced, breakable,
  colback=black!3, colframe=black!55, boxrule=0.5pt, arc=1.5pt,
  left=5pt, right=5pt, top=4pt, bottom=4pt,
  fonttitle=\bfseries\small, coltitle=black, colbacktitle=black!10,
  title={#2}, fontupper=\small,
  subtitle style={before skip=6pt, after skip=4pt, colback=black!10, fonttitle=\bfseries\small},
  #1}

\title{\model{}: Bidirectionally Coupling Time Series Foundation Models and LLMs for Multimodal Reasoning and Forecasting}

\author{Ziyang Zhang$^{1,2}$ \quad Shenyi Li$^{2}$ \quad Wen Wu$^{2}$ \quad Chao Zhang$^{1,2}$\thanks{Corresponding author.} \\
\textnormal{$^{1}$Tsinghua University \quad $^{2}$Shanghai Artificial Intelligence Laboratory} \\
\textnormal{\texttt{ziyang-z24@mails.tsinghua.edu.cn}, \texttt{cz277@tsinghua.edu.cn}} \\
\textnormal{\texttt{\{lishenyi, wuwen\}@pjlab.org.cn}}
}
\hypersetup{colorlinks=true, linkcolor={red!60!black}, citecolor={blue!60!black}, urlcolor={magenta!60!black}, pdftitle={TimeOmni-o1: Bidirectionally Coupling Time Series Foundation Models and LLMs for Multimodal Reasoning and Forecasting}, pdfauthor={Ziyang Zhang, Shenyi Li, Wen Wu, Chao Zhang}}

\iclrfinalcopy

\begin{document}

\maketitle
\lhead{Preprint}

\begin{abstract}
Time series from real-world systems are often accompanied by text, such as a report on an event that affects them. Two common tasks on these series are understanding, \textit{e.g.}, answering questions about a series, and forecasting its future values. Both tasks require reasoning over the series and the text together. Moreover, understanding outputs text while forecasting outputs a time series, so serving both requires a model that reads and generates both modalities. Large language models (LLMs) read time series as digit tokens and handle long series poorly. Time series foundation models (TSFMs) forecast well from large-scale pretraining but typically neither read nor generate text. Time series language models add a time series modality to the LLM but forecast on the LLM side, leaving the TSFM's pretrained forecasting ability unused. We present \textbf{\model{}}, which couples a pretrained TSFM and an LLM bidirectionally. In one direction, the TSFM encodes the series in compressed form for the LLM, which reasons over it together with the text. In the other, the LLM's reasoning is fed back into the TSFM, which produces the forecast conditioned on it. One model architecture thus reads and generates both series and text, reasons in text, and draws on the TSFM's pretrained forecasting ability. To train it, we build \textbf{\dataset{}}, a multimodal time series reasoning dataset with annotated reasoning traces, covering both tasks across 12 domains with 500k+ samples. On \dataset{}, \model{} outperforms existing models on both understanding and forecasting across most domains, as well as the LLM and the TSFM inside it when used individually. The gains carry over to three open benchmarks, where it outperforms the dedicated time series models.\footnote{Code: \url{https://github.com/OpenTSLab/TimeOmni-o1}. Project page: \url{https://windowso.github.io/TimeOmni-o1-web/}.}
\end{abstract}

\section{Introduction}
\label{sec:intro}

Many real-world domains such as energy, transportation, finance, and healthcare rely on time series, which typically involve two tasks: understanding a series, \textit{e.g.}, answering questions about it, and forecasting its future values~\citep{timemmd,scits}. In both tasks, the time series is often accompanied by text, such as a report on an event that affects it, and this text often carries information the numbers do not~\citep{cik,mtbench}. Both tasks therefore call for reasoning across the two modalities: the output cannot be read off the numbers alone, but must be derived from what the series shows and what the text says. The two tasks differ in output: understanding outputs text, whereas forecasting outputs a time series. Serving both tasks thus requires a model that can not only read but also generate both text and time series.

Recent work applies large language models (LLMs) to time series tasks, drawing on their ability to reason in text~\citep{llmtime,lstprompt}. However, an LLM reads a time series as digit tokens and handles it poorly once the series is long~\citep{scits}. Time series foundation models (TSFMs) are pretrained on large-scale time series and thus have zero-shot forecasting ability~\citep{chronos,timesfm,moirai}, but typically neither read nor generate text. Several studies build time series language models (TSLMs) by adding a time series modality to the LLM, either as discrete tokens~\citep{timeomni1,timer1} or as encoder features~\citep{chattsllm,opentslm,tsllm}. However, TSLMs produce the forecast on the LLM side, as digit tokens or from a prediction head, and leave a TSFM's pretrained forecasting ability unused. To our knowledge, none of these three types of models reads and generates both series and text, reasons in text, and still draws on a TSFM's pretrained forecasting ability at once.

To address this gap, we present \textbf{\model{}}, one model architecture that reads and generates both series and text, reasons in text, and draws on a pretrained TSFM's forecasting ability. Its key idea is to couple the TSFM and an LLM bidirectionally, so that each feeds the other. In one direction, the TSFM encodes the time series and passes it to the LLM in compressed form, so the LLM reasons over the series together with textual context. For understanding, the LLM then generates the final answer directly. In the other direction, for forecasting, the reasoning is fed back into the TSFM, which produces the forecast conditioned on it.

Training \model{} requires data that pair time series with text, cover both understanding and forecasting, and carry annotated reasoning traces. Few existing datasets provide all three at scale~\citep{timemmd,streasoner}. We therefore build \textbf{\dataset{}}, a multimodal time series reasoning dataset that meets these requirements. It collects public multimodal time series corpora under one schema and adds reasoning traces to some samples. It holds 502{,}586 training and 71{,}067 test samples, spanning 12 domains.

We benchmark \model{} and baselines on \dataset{}. \model{} outperforms the baselines on both understanding and forecasting across most domains, and also outperforms the LLM and the TSFM inside it when used individually, so the gain comes from the bidirectional coupling rather than from either component by itself. The gains carry over to three open benchmarks~\citep{scits,streasoner,cik}, where it outperforms the dedicated time series models each benchmark reports.

Our contributions are as follows: \textbf{1)} We propose \model{}, which bidirectionally couples a TSFM and an LLM for multimodal time series understanding and forecasting. One model architecture thus reads and generates both series and text, reasons in text, and draws on the TSFM's pretrained forecasting ability. \textbf{2)} We build \dataset{}, a multimodal time series reasoning dataset that covers both understanding and forecasting tasks across 12 domains, with annotated reasoning traces. \textbf{3)} \model{} outperforms existing models on both understanding and forecasting across most domains on \dataset{}. The gains carry over to three open benchmarks, where it outperforms the dedicated time series models each benchmark reports.

\section{Related Work}
\label{sec:related}

\textbf{Coupling TSFMs with LLMs.}
Most existing couplings connect the two models in one direction and serve a single task. In the LLM-to-TSFM direction, raw text~\citep{doublecast}, instructions written by an LLM~\citep{chronosteer}, or reasoning traces~\citep{stride} are injected into a pretrained TSFM to condition its forecast. All of these are forecast-only. In the TSFM-to-LLM direction, TS-Reasoner~\citep{tsreasoner} aligns a frozen TSFM into an LLM for understanding and reasoning, but the TSFM only feeds the LLM and never produces a forecast. Unlike either line of work, \model{} couples the TSFM and the LLM bidirectionally: the TSFM feeds the LLM, and the LLM's reasoning is fed back into the TSFM.

\textbf{Unified time series understanding and forecasting.}
Several recent models unify understanding and forecasting in one model \citep{tsllm,units,timemqa}. Most build on an LLM, which reads the time series as digit tokens~\citep{chattime,timemqa}, as a rendered image, or through an encoder without pretraining~\citep{opentslm,scits}. In all of them, the forecast is produced on the LLM side, as decoded digit tokens, through a prediction head, or as a rendered plot~\citep{timeomnivl}, so a pretrained TSFM's forecasting ability goes unused. UniTS~\citep{units} instead unifies the two tasks within the time series modality alone and has no text interface. \model{} instead reads and generates both series and text through one architecture, and forecasts through a TSFM, so it draws on the TSFM's pretrained forecasting ability.

\textbf{Multimodal time series reasoning datasets.}
On the understanding task, existing datasets with reasoning traces each cover a limited range of tasks and domains: the OpenTSLM chain-of-thought sets~\citep{opentslm} cover physiological signals alone, TelecomTS~\citep{telecomts} a single domain, 5G network operations, and RATs40K~\citep{timera} a single task, anomaly diagnosis, while TSRBench~\citep{veritime} and HiTSR~\citep{hitsr} rely mostly on synthetic series and pose only question-answering tasks. ST-Bench~\citep{streasoner} does include forecasting, but only on synthetic spatio-temporal series. On the forecasting task, context-aided benchmarks supply event text but no reasoning traces~\citep{cik,chattime,fidelts,mtbench,newsforecast,timemmd}. \dataset{} collects the six reasoning-annotated corpora above under one schema and annotates reasoning traces for forecasting data, so it covers both understanding and forecasting tasks with annotated reasoning traces. Table~\ref{tab:dataset-cmp} compares \dataset{} with the closest of these corpora.

\begin{figure}[t]
\begin{center}
\includegraphics[width=\linewidth]{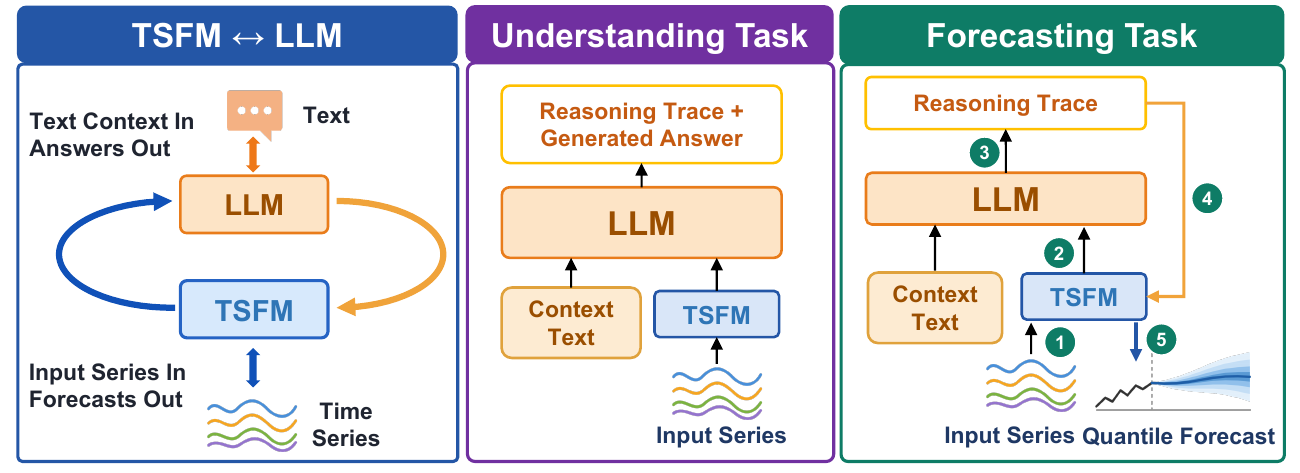}
\end{center}
\caption{Overview of \model{}. Left: a pretrained TSFM and an LLM are coupled bidirectionally. The TSFM is the interface for time series, taking the input series and producing the forecast, and the LLM is the interface for text, taking the context and producing the reasoning trace and the answer. Middle: for understanding, the LLM reasons over the TSFM-encoded time series and the context text, and generates the reasoning trace and the answer. Right: for forecasting, the time series enters through the TSFM (1, 2) and the LLM generates the reasoning trace (3). The LLM's hidden states over the whole sequence, holding the time series, the context, and the trace, are then fed back into the TSFM (4), which generates the quantile forecast (5).}
\label{fig:overview}
\vspace{-10pt}
\end{figure}

\section{TimeOmni-o1}
\label{sec:model}

\subsection{Overview}
\label{sec:workflow}

\model{} performs both the understanding and the forecasting task with one architecture (Figure~\ref{fig:overview}). The input is a multivariate time series of $C$ channels, paired with a textual context: a question, background, or an event description. For understanding, the model generates a reasoning trace followed by a textual answer. For forecasting, the trace instead ends in a \emph{forecast instruction}, and the output is a quantile forecast over the requested horizon, the window of future steps each sample specifies. The instruction states the window over which the event is expected to act, and the direction, shape and relative magnitude of its effect within that window.

\begin{figure}[t]
\begin{center}
\includegraphics[width=\linewidth]{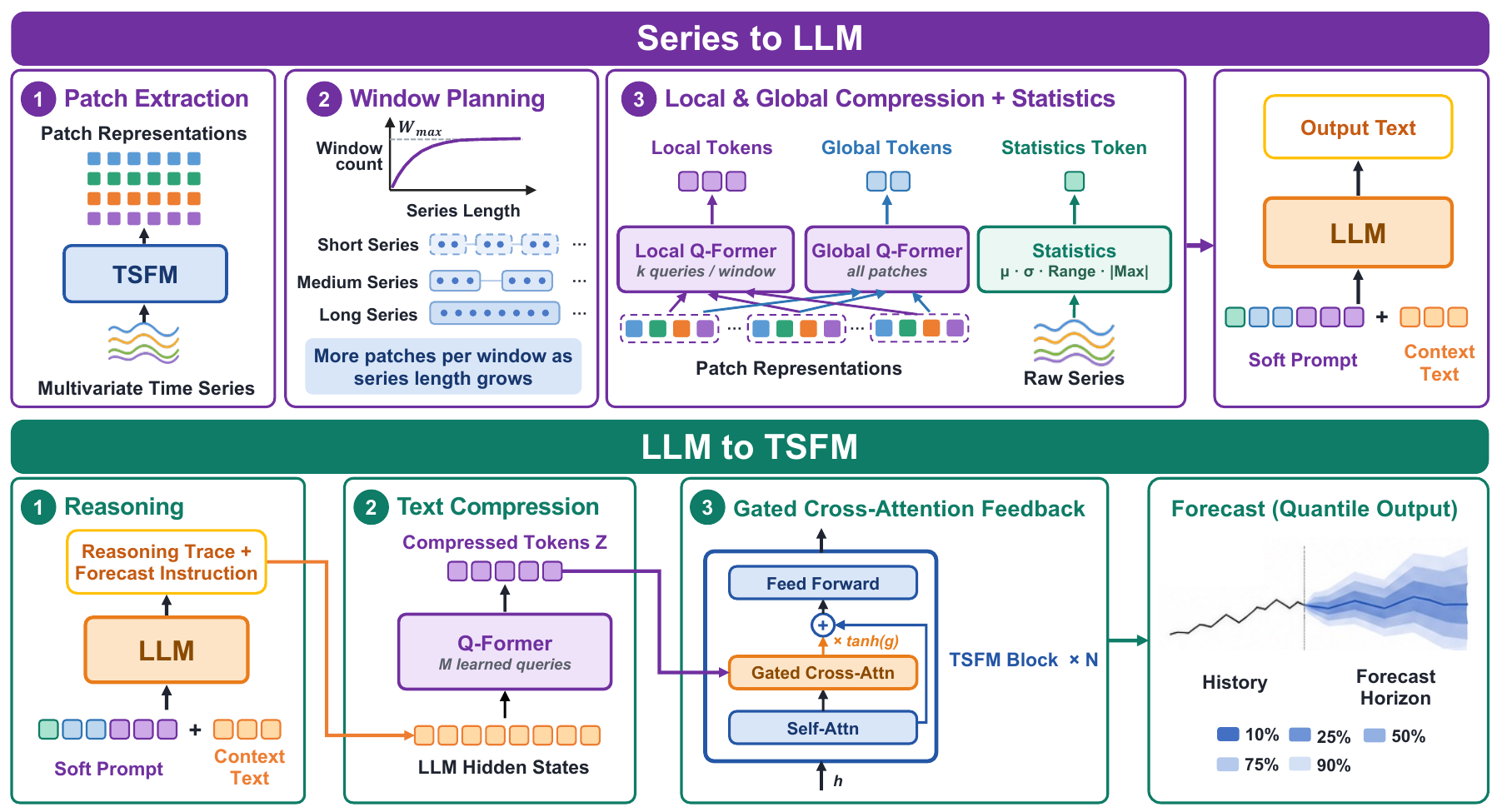}
\end{center}
\caption{The two pathways of \model{}. Top, time series to LLM (\S\ref{sec:series2llm}): the TSFM maps the time series into patch representations (1). A planning rule sets the window count from the time series length (2). The Q-Former compresses the patches into local and global tokens; with the statistics token they form the soft prompt (3), which the LLM reads alongside the context text. Bottom, LLM to TSFM (\S\ref{sec:llm2tsfm}): the LLM generates the reasoning trace and the forecast instruction (1). The feedback Q-Former compresses the LLM hidden states into $M$ tokens $\mathbf{Z}$ (2). Gated cross-attention injects $\mathbf{Z}$ into every TSFM block (3), and the TSFM generates the quantile forecast.}
\label{fig:arch}
\vspace{-10pt}
\end{figure}

Both task types share one workflow in which each component performs the function it was pretrained for: the LLM generates all text, and the TSFM encodes the time series and produces the forecast. The two components are coupled bidirectionally, with one pathway in each direction. On the series-to-LLM pathway, the TSFM maps the time series to patch representations, which are compressed into a bounded soft prompt that the LLM reads alongside the context to generate the trace and the answer or instruction. On the LLM-to-TSFM pathway, used for forecasting, the LLM's hidden states are compressed and injected into the TSFM, which produces the forecast conditioned on them.

\subsection{Series to LLM}
\label{sec:series2llm}
To process long time series, the series-to-LLM pathway compresses the TSFM's patch representations into a bounded soft prompt of local, global, and statistics tokens, which the LLM reads alongside the context (Figure~\ref{fig:arch}, top). The TSFM produces $P$ patch representations per channel, and the channels of one sample interact through the TSFM's cross-channel attention, giving $N=P\times C$ patch representations in total. A sliding-window Q-Former then compresses them. It partitions the patch sequence into $W$ windows and summarizes each window into $k$ local tokens with learned queries shared by all windows. The window count is set as $W=\min\bigl(W_{\min}+b\log_2 N,\,W_{\max}\bigr)$, which grows logarithmically with $N$ and is capped at $W_{\max}$. Since $W$ grows only logarithmically while $N$ grows linearly with the series length, the window width $\lceil N/W\rceil$ increases as the series lengthens, and once $W$ reaches $W_{\max}$, further length is absorbed entirely by wider windows (Appendix~\ref{app:swqformer}). The local tokens alone may lose two kinds of information. First, the Q-Former summarizes each window independently, and the TSFM itself attends only within a chunk of bounded length, so global structure would have to be recovered by the LLM. A small set of global queries therefore attends to all $P\times C$ patches in a single cross-attention layer, and their outputs are prepended to the local tokens. Second, each channel is normalized by a single $(\mu,\sigma)$ pair over its whole history, so a small fluctuation and the same shape at ten times its amplitude yield identical representations. A statistics token restores the scale: each channel's $(\mu,\sigma)$, range and peak amplitude are projected into the LLM input space and placed at the head of the soft prompt.

\subsection{LLM to TSFM}
\label{sec:llm2tsfm}
The LLM-to-TSFM pathway compresses the LLM's hidden states and injects them into every TSFM block through gated cross-attention, so the forecast is conditioned on the reasoning (Figure~\ref{fig:arch}, bottom). Only forecasting uses this pathway. For understanding, the LLM generates the answer itself after reasoning. For forecasting, the TSFM produces the output, so the reasoning trace must be passed to it. The trace ends in the forecast instruction of Section~\ref{sec:workflow}. A feedback Q-Former with $M$ learned queries cross-attends to the LLM's last-layer hidden states over the soft prompt, the context and the trace, and produces $\mathbf{Z}\in\mathbb{R}^{M\times d}$, with $d$ the TSFM's hidden size. In block $\ell$ of the TSFM, the gated cross-attention updates the patch representations $\mathbf{h}_\ell$ as
\begin{equation}
\mathbf{h}_\ell \leftarrow \mathbf{h}_\ell + \tanh(g_\ell)\cdot \mathrm{CrossAttn}_\ell\bigl(\mathbf{h}_\ell,\ \mathbf{Z}\bigr),
\end{equation}
where $g_\ell$ is a scalar gate. The reasoning thus changes the TSFM's internal representations rather than modifying the output forecast directly, and the quantiles are still generated by the TSFM.

\section{\dataset{}}
\label{sec:dataset}

\begin{table}[t]
\centering
\footnotesize
\setlength{\tabcolsep}{3pt}
\caption{\dataset{} against related multimodal time series corpora. The full comparison, including all eleven source corpora of \dataset{}, is Table~\ref{tab:app-dataset-cmp} in Appendix~\ref{app:related-tables}. \textbf{Und.}~/~\textbf{Fc.}: the corpus poses understanding / forecasting tasks. \textbf{Trace}: every sample carries an annotated reasoning trace; \emph{partial}: only a subset does (2.3k of TSR-Suite's 23.6k samples). \textbf{Domains}: as counted by the source. \textbf{Series}: real, synthetic, or both. \textbf{Train split}: the corpus has a training split, rather than being an evaluation-only benchmark. \textbf{Samples}: total instances as reported by the source.}
\label{tab:dataset-cmp}
\begin{tabular}{lccclccr}
\toprule
Dataset & Und. & Fc. & Trace & Domains & Series & Train split & Samples \\
\midrule
OpenTSLM CoT sets \citep{opentslm} & \cmark & \xmark & \cmark & 1 & real & \cmark & 326k \\
RATs40K \citep{timera} & \cmark & \xmark & \cmark & 10 & real & \cmark & 39.6k \\
ST-Bench \citep{streasoner} & \cmark & \cmark & \cmark & 10 & synthetic & \cmark & 28.2k \\
TSR-Suite \citep{timeomni1} & \cmark & \cmark & partial & 10 & real & \cmark & 23.6k \\
TSQA \citep{timemqa} & \cmark & \cmark & \xmark & 12 & both & \cmark & 193k \\
SciTS \citep{scits} & \cmark & \cmark & \xmark & 12 & both & \cmark & 52.1k \\
Time-MMD \citep{timemmd} & \xmark & \cmark & \xmark & 9 & real & \cmark & 17.1k \\
CiK \citep{cik} & \xmark & \cmark & \xmark & 7 & both & \xmark & 355 \\
\midrule
\textbf{\dataset{} (Ours)} & \cmark & \cmark & \cmark & 12 & both & \cmark & 574k \\
\bottomrule
\end{tabular}
\vspace{-6pt}
\end{table}

Table~\ref{tab:dataset-cmp} collects the related multimodal time series corpora, and no corpus in it combines a reasoning trace on every sample, both task types and real-world time series. We therefore build \dataset{}, a multimodal time series reasoning dataset with annotated reasoning traces, covering both the understanding and the forecasting task. It holds 573{,}653 samples: 568{,}586 for the understanding task over 12 domains and 5{,}067 for the event-conditioned forecasting task over 5 domains.

\subsection{Understanding Task}
The understanding task is built from six public reasoning-annotated corpora: the OpenTSLM chain-of-thought sets \citep{opentslm}, ST-Bench \citep{streasoner}, RATs40K \citep{timera}, TSRBench \citep{veritime}, TelecomTS \citep{telecomts} and HiTSR \citep{hitsr}. They are unified into one schema of time series, question, reasoning trace and answer. The questions cover open question answering, multiple choice, classification, anomaly detection and diagnosis, and multi-step numerical calculation. The samples span 12 domains: physiology, telecommunications, urbanism, information technology, neuroscience, earth science, meteorology, economics, bioacoustics, manufacturing, energy, and synthetic data.

\subsection{Forecasting Task}
\label{sec:dataset-fc}
The forecasting task is event-conditioned. Each sample pairs a time series history with a background and the description of an event (\textit{e.g.}, a heatwave alert, a holiday, a market event), and asks for a forecast over a horizon. The time series and their accompanying texts are collected from five public multimodal forecasting sources: CGTSF \citep{chattime}, Fidel-TS \citep{fidelts}, MTBench \citep{mtbench}, News Forecast \citep{newsforecast} and Time-MMD \citep{timemmd}. These sources contribute ten subsets in five domains: solar power, grid load, traffic flow, finance and climate. They carry no reasoning annotations, so we annotate every sample with Claude Sonnet 4.6\footnote{\url{https://www.anthropic.com/news/claude-sonnet-4-6}} to add an event description condensed from the accompanying text, a reasoning trace, and a forecast instruction. The annotator writes the reasoning trace and the forecast instruction from the history and the event while seeing the future values, so the instruction matches the realized future, but it is prompted not to cite any specific future value. The window stated in the instruction, over which the event is expected to act, is the event region of interest (ROI). We record it as a separate field in each sample for ROI-weighted training and ROI evaluation. The subsets and the annotation protocol are detailed in Appendix~\ref{app:dataset}.

\subsection{Annotation Quality Review}
\label{sec:dataset-review}
Trace quality is checked by human review. Two independent reviewers rate 1{,}000 traces, sampled with stratification by task and domain, against the input time series, the context, and the annotated answer. A trace fails if its reasoning contradicts the time series, the context, or the final answer, or if it contains a factual error. On average, 93.3\% of the understanding traces and 97.7\% of the forecasting traces pass, and the two reviewers give the same pass/fail label for 96.6\% of the traces. The review protocol and the per-task pass rates are given in Appendix~\ref{app:audit}.

\section{Experimental Setup}
\label{sec:setup}

\textbf{Model and training.} The TSFM of \model{} is instantiated with Chronos-2 \citep{chronos2}, an encoder-only patch-based multivariate forecaster with native quantile output, and its LLM with Qwen3.5-9B \citep{qwen3}. Training supervises the trace and the answer or instruction with text cross-entropy loss, and the forecast with a quantile loss reweighted inside the ROI. The workflow is differentiable end to end, so the quantile loss back-propagates to the LLM through the LLM-to-TSFM pathway and the forecast error also trains the reasoning. The LLM is adapted with LoRA \citep{lora}, while the TSFM and Q-Formers are fully trained. Appendix~\ref{app:hyper} gives the training details, and Appendix~\ref{app:comp-swap} the results with another TSFM and another LLM.

\textbf{Baselines.} We evaluate the following models on \dataset{}: ChatTime-1-7B \citep{chattime}, Chat-TS-8B \citep{chattsllm}, Time-MQA \citep{timemqa}, TimeOmni-VL-15B \citep{timeomnivl}, TimeOmni-1-4B \citep{timeomni1}, TimeReasoner \citep{timereasoner}, DoubleCast \citep{doublecast}, TimeOmni \citep{scits}, UniTS \citep{units}, and the LLM and the TSFM components of \model{}. The LLM component reads and generates the time series as full-precision text under the same token budget as \model{}. The TSFM component, which has no text interface, is fine-tuned with per-subtask classification heads on the understanding subtasks it supports. The LLM and the TSFM components of \model{} are evaluated both zero-shot and fine-tuned on the \dataset{} training split, where they reuse \model{}'s training configuration. TimeOmni and UniTS are fine-tuned on the same split and follow their official fine-tuning recipes. The remaining models are evaluated zero-shot on the tasks they support.

\textbf{Metrics.} For understanding, each subtask is scored by its source's native protocol, and subtask scores aggregate into the 12 domains by sample count. For forecasting, we report CRPS over the full horizon, weighted by each model's parse success rate.

\section{Results and Discussion}
\label{sec:results}

\subsection{Main Results}
\label{sec:results-main}

\begin{table}[t]
\centering
\footnotesize
\setlength{\tabcolsep}{1pt}
\caption{\textbf{Understanding} on the \dataset{} test set. \textbf{Avg.} is the mean of the 12 domain scores. Every subtask is scored by its source's native protocol. Bold marks the best score in each column. $\dagger$: fine-tuned on the \dataset{} training split. \model{}'s TSFM is fine-tuned with per-subtask classification heads; it has no text interface and can attempt only 11 of the 21 subtasks (18\% of test samples). A parenthesized cell gives the share of that domain's test samples it can attempt; a score is reported only where this share is 100\%, so Avg. is undefined.}
\label{tab:main-und}
\begin{tabular}{lccccccccccccc}
\toprule
Model & Phys. & Tele. & Synth. & Urban. & IT & Neuro. & Earth. & Meteo. & Econ. & Bioac. & Manuf. & Ener. & \textbf{Avg.} \\
\midrule
\multicolumn{14}{l}{\emph{Time series models (zero-shot / fine-tuned$^\dagger$)}} \\
ChatTime-1-7B & 0.9 & 0.0 & 0.0 & 0.1 & 15.3 & 0.0 & 21.2 & 31.7 & 8.4 & 0.0 & 20.8 & 0.0 & 8.2 \\
Chat-TS-8B & 0.8 & 0.0 & 0.0 & 0.1 & 22.0 & 0.0 & 11.5 & 3.4 & 9.5 & 0.0 & 20.3 & 0.0 & 5.6 \\
Time-MQA-7B & 1.7 & 0.0 & 1.4 & 0.3 & 22.5 & 25.5 & 30.5 & 31.8 & 9.5 & 0.0 & 22.9 & 0.0 & 12.2 \\
TimeOmni-VL-15B & 10.6 & 10.7 & 0.8 & 37.2 & 26.0 & 28.6 & 28.5 & 32.0 & 13.2 & 0.0 & 25.3 & 6.3 & 18.3 \\
TimeOmni-1-4B & 27.4 & 18.5 & 71.0 & 44.5 & 50.6 & 50.3 & 37.8 & 46.9 & 55.8 & 0.0 & 37.3 & 7.5 & 37.3 \\
TimeOmni$^\dagger$ & 58.9 & \textbf{97.6} & 93.8 & 95.7 & 77.4 & 90.0 & 46.7 & 94.5 & 62.8 & 77.0 & 40.9 & 61.9 & 74.8 \\
\midrule
\multicolumn{14}{l}{\emph{Per component of \model{} (zero-shot / fine-tuned$^\dagger$)}} \\
\model{}'s LLM & 1.3 & 0.2 & 0.1 & 0.2 & 15.7 & 7.5 & 21.3 & 32.5 & 12.5 & 0.0 & 21.1 & 1.3 & 9.5 \\
\model{}'s LLM$^\dagger$ & 53.5 & 96.9 & \textbf{98.5} & 79.3 & 80.1 & 93.8 & 56.1 & \textbf{97.1} & 72.0 & 65.0 & 52.0 & 54.4 & 74.9 \\
\model{}'s TSFM$^\dagger$ & \pct{19} & \pct{0} & \pct{0.3} & \pct{1} & \pct{91} & 97.8 & \pct{95} & \pct{91} & \pct{56} & 77.5 & \pct{93} & \pct{94} & -- \\
\midrule
\textbf{\model{} (Ours)} & \textbf{65.0} & 97.0 & 97.4 & \textbf{96.2} & \textbf{84.1} & \textbf{98.5} & \textbf{73.8} & 97.0 & \textbf{80.6} & \textbf{78.5} & \textbf{64.6} & \textbf{78.1} & \textbf{84.2} \\
\bottomrule
\end{tabular}
\vspace{-10pt}
\end{table}

\begin{table}[t]
\centering
\small
\setlength{\abovecaptionskip}{3pt}
\caption{\textbf{Forecasting} on the \dataset{} test set: CRPS over the full horizon, lower is better. \textbf{Avg.} is the mean of the five domains; scores of models with parse failures are success-rate weighted. Seasonal naive copies the last horizon window of the history. Bold marks the best score in each column. $\dagger$: fine-tuned on the \dataset{} training split.}
\label{tab:main-fc}
\begin{tabular}{lcccccc}
\toprule
Model & Solar & Load & Traffic & Finance & Climate & \textbf{Avg.} \\
\midrule
\multicolumn{7}{l}{\emph{Naive reference}} \\
Seasonal naive & 0.717 & 0.063 & 0.378 & 0.051 & 0.126 & 0.267 \\
\midrule
\multicolumn{7}{l}{\emph{Time series models (zero-shot / fine-tuned$^\dagger$)}} \\
ChatTime-1-7B & 0.898 & 0.106 & 0.376 & 0.058 & 0.107 & 0.309 \\
Time-MQA-7B & 2.868 & 0.362 & 0.807 & 0.162 & 0.116 & 0.863 \\
TimeReasoner & 0.640 & 0.081 & 0.271 & 0.041 & 0.110 & 0.229 \\
DoubleCast & 0.530 & 0.083 & 0.180 & 0.029 & 0.065 & 0.177 \\
TimeOmni-VL-15B & 4.597 & 0.466 & 1.305 & 0.101 & 0.218 & 1.337 \\
TimeOmni-1-4B & 1.000 & 0.257 & 0.631 & 0.057 & 0.132 & 0.415 \\
UniTS-x128$^\dagger$ & 0.865 & 0.059 & 0.276 & 0.033 & 0.111 & 0.269 \\
TimeOmni$^\dagger$ & 0.468 & 0.042 & 0.224 & 0.033 & 0.085 & 0.170 \\
\midrule
\multicolumn{7}{l}{\emph{Per component of \model{} (zero-shot / fine-tuned$^\dagger$)}} \\
\model{}'s LLM & 2.277 & 0.322 & 0.622 & 0.190 & 0.217 & 0.726 \\
\model{}'s LLM$^\dagger$ & 1.113 & 0.071 & 0.212 & 0.047 & 0.097 & 0.308 \\
\model{}'s TSFM & 0.511 & 0.066 & 0.187 & 0.032 & 0.065 & 0.172 \\
\model{}'s TSFM$^\dagger$ & 0.378 & 0.037 & 0.138 & 0.025 & 0.065 & 0.129 \\
\midrule
\textbf{\model{} (Ours)} & \textbf{0.357} & \textbf{0.032} & \textbf{0.113} & \textbf{0.022} & \textbf{0.056} & \textbf{0.116} \\
\bottomrule
\end{tabular}
\vspace{-16pt}
\end{table}

\textbf{Understanding.} On the understanding task of \dataset{} (Table~\ref{tab:main-und}), \model{} averages 84.2 and leads all six time series models on 11 of the 12 domains. The outputs of these models fall into three types. The text that ChatTime and Chat-TS generate contains no answer to the question, since they continue the time series or echo a template phrase unrelated to the question. Time-MQA and TimeOmni-VL read the time series and generate an analysis, but it usually ends without an answer. TimeOmni-1 and the fine-tuned TimeOmni answer in the expected form, so their scores measure content. TimeOmni fine-tuned on \dataset{} scores high where the answer template is fixed and easy to guess, and low where the question requires understanding the time series.

Comparing \model{} with its components used individually shows what the coupling adds. The LLM in the zero-shot setting generates a long analysis that reaches no answer within the budget on most questions. The fine-tuned LLM exceeds \model{} on Synth.\ and Meteo.\ because their questions ask for exact values of a short time series. A time series read as text keeps every value exactly, whereas the soft prompt encodes the series only approximately. On the other ten domains the LLM falls behind, because a long time series written out as digits easily exceeds the context budget. Even when the time series fits the budget, full-precision digits may draw attention to individual values at the expense of trend, periodicity and anomaly morphology, which the soft prompt presents directly. The TSFM has no text interface and attempts only the closed-label subtasks (Table~\ref{tab:main-und}).

\textbf{Forecasting.} On the forecasting task (Table~\ref{tab:main-fc}), \model{} has the lowest CRPS on every domain. A seasonal-naive forecast that copies the last horizon window of the history outperforms five of the eight time series models and both LLM rows. ChatTime, Time-MQA and TimeOmni-1 generate the numbers as text and often yield no valid forecast, because the output is unparseable, constant or on a wrong scale. TimeOmni-VL reads the time series back from a rendered plot and therefore loses the absolute level. TimeReasoner, UniTS, DoubleCast and TimeOmni produce valid forecasts, but none outperforms \model{}'s fine-tuned TSFM component.

Both components of \model{}, used individually, trail \model{} on every forecasting domain. The LLM generates the forecast as text and fails like ChatTime, Time-MQA and TimeOmni-1. Under half of its zero-shot forecasts parse, and most failures are non-terminating digit sequences. Fine-tuning makes most of the LLM's forecasts parse, but even then the LLM forecasts worse than the zero-shot TSFM. The TSFM improves with fine-tuning but still trails \model{} on every domain, by about 10\% of its average CRPS. The two rows share the forecaster and the training time series and differ only in whether the event text is passed to the forecaster, so their gap is what the text contributes to the forecast.

\subsection{Quality of the Generated Reasoning}
\label{sec:results-trace}

Accuracy alone does not justify the traces, because a plausible trace may be unrelated to the answer. The generated reasoning is therefore scored by an LLM judge (GPT-5.4\footnote{\url{https://developers.openai.com/api/docs/models/gpt-5.4}}) on two metrics. \textit{Consistency} compares the generated reasoning with the annotated trace of the same sample. \textit{Supportiveness} checks whether the reasoning supports the model's own output. Each trace is scored twice, once matched against its own sample and once mismatched against a random other sample's annotation or output, which gives the chance level.

\setlength{\intextsep}{2pt}
\begin{wraptable}[8]{r}{0.58\textwidth}
\centering
\footnotesize
\setlength{\abovecaptionskip}{0pt}
\setlength{\tabcolsep}{2pt}
\vspace{-10pt}
\caption{Reasoning-trace quality, scored 1 to 5 by an LLM judge, GPT-5.4. 
}
\label{tab:trace-eval}
\begin{tabular}{lcccc}
\toprule
& \multicolumn{2}{c}{Understanding} & \multicolumn{2}{c}{Forecasting} \\
\cmidrule(lr){2-3}\cmidrule(lr){4-5}
Metric & Matched & Mismatched & Matched & Mismatched \\
\midrule
Consistency & 4.24 & 1.89 & 3.18 & 2.17 \\
Supportiveness & 4.72 & 1.45 & 3.26 & 2.53 \\
\bottomrule
\end{tabular}
\end{wraptable}
\setlength{\intextsep}{12pt plus 2pt minus 2pt}
Table~\ref{tab:trace-eval} reports both metrics over 844 understanding samples stratified by domain and all forecasting test samples. On both metrics and both tasks, the matched score is well above the mismatched score. The traces thus agree with the annotations and support the model's own output, and they are specific to their sample rather than generic text that fits any time series.

A high judge score still leaves open whether the output follows the trace or the trace rationalizes the output post hoc, so this is tested separately on each task. On understanding, the matched supportiveness of Table~\ref{tab:trace-eval} (4.72) is split by whether the model's answer is correct. A post-hoc rationalization would support wrong answers as strongly as correct ones. The observed score is lower on wrong answers (4.43 against 4.90) in all 12 domains, which suggests that the answer depends on the trace. On forecasting, the instruction's statement of relative magnitude is replaced in turn by five statements, ordered from the lowest to the highest magnitude, and the forecast is regenerated. If the forecast depends on the instruction, a higher stated magnitude should yield a higher forecast. The quantile forecasts follow the edit with a rank correlation of 0.74 and 97.0\% direction agreement, against chance levels of zero and 50\%. The forecast instruction is an input the forecaster uses, not a description written alongside the forecast. The full protocol is given in Appendix~\ref{app:trace-eval}.

\subsection{When Does Reasoning Help?}
\label{sec:results-ablation}

\begin{figure}[t]
\centering
\includegraphics[width=0.85\linewidth]{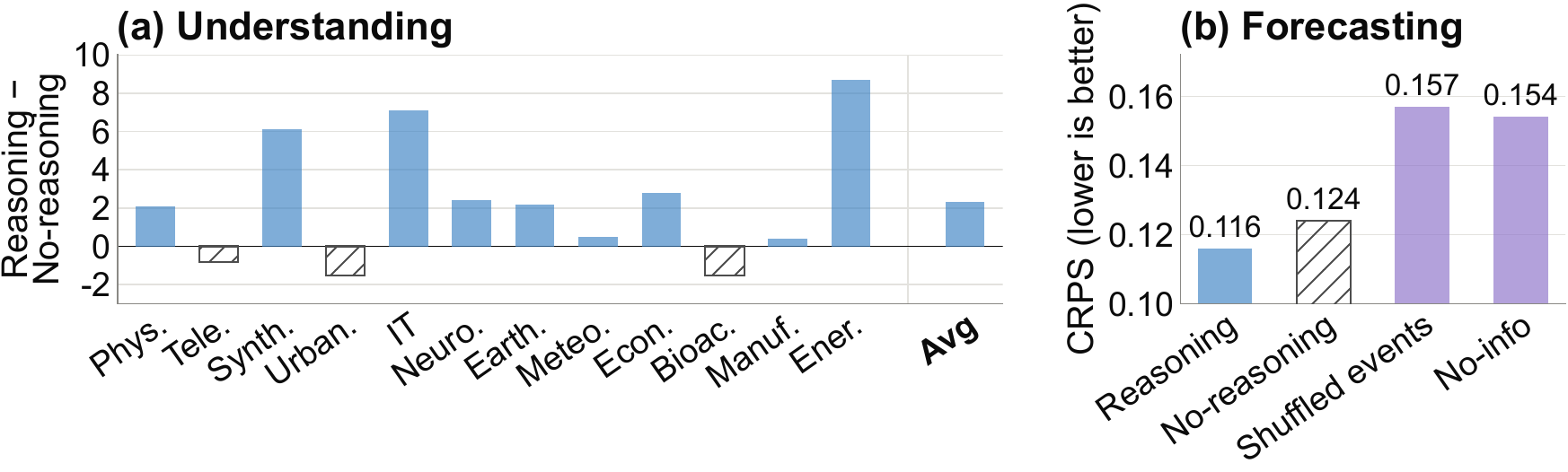}
\vspace{-6pt}
\caption{Reasoning against no-reasoning, the identical recipe trained without reasoning traces. \textbf{(a)} Understanding score difference (reasoning minus no-reasoning) per domain. \textbf{(b)} Shuffled events and no-info replace the event text by another sample's event or by an uninformative placeholder.}
\label{fig:abl}
\vspace{-12pt}
\end{figure}

Figure~\ref{fig:abl} compares \model{} with its no-reasoning counterpart. On understanding, reasoning scores above no-reasoning on nine of the 12 domains and below it on three. In the cases examined, the gains on the nine domains come from questions answered in several steps. On counting questions, the trace counts the events one by one, whereas the no-reasoning model states a count directly and often miscounts (Appendix~\ref{app:case}, case U1). On anomaly classification, the trace checks each class's defining features, whereas the no-reasoning model often labels a normal time series as an anomaly (case U2). The three losses (Tele., Urban., Bioac.) occur on questions answered by a single judgment, where the trace has no intermediate steps to add. On forecasting, reasoning lowers the average CRPS by about 6\%. The trace derives the shape from the event and weighs conflicting events to decide the direction, whereas the no-reasoning model mainly follows a magnitude or time span that the event text states directly (cases F1, F2). Both test-time replacements of the event text raise CRPS (Figure~\ref{fig:abl}(b)), which shows that the forecast depends on the correct event text.

\subsection{Validation on Additional Benchmarks}
\label{sec:results-ood}

\begin{figure}[t]
\centering
\includegraphics[width=\linewidth]{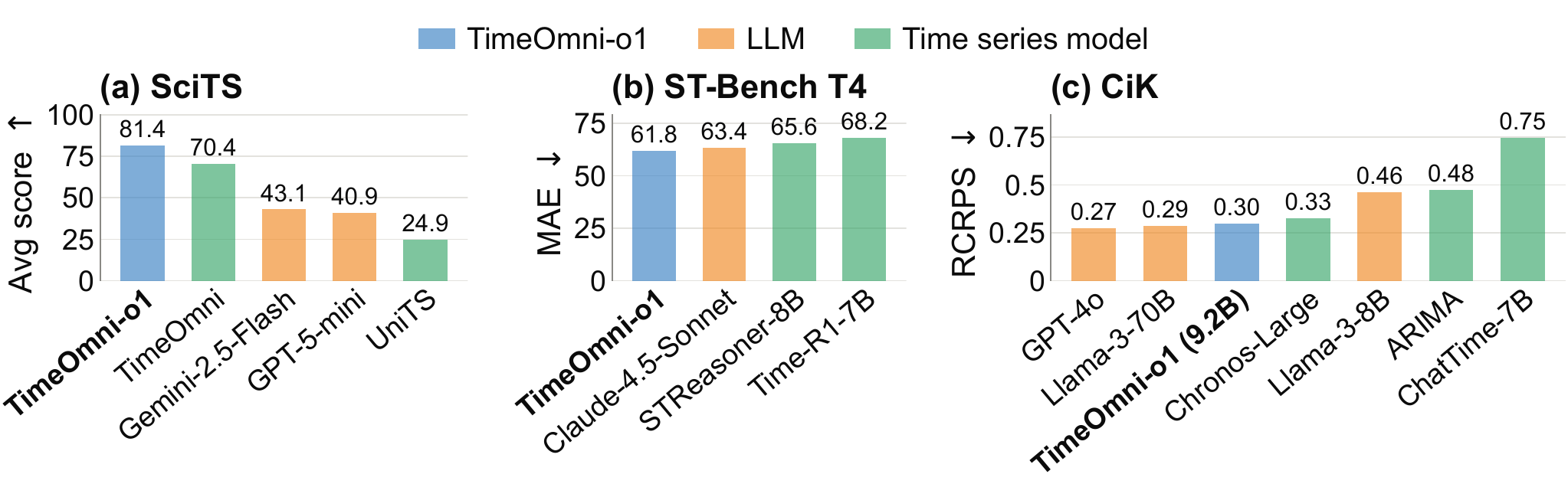}
\vspace{-14pt}
\caption{Validation on additional benchmarks. \textbf{(a)} SciTS understanding, mean per-domain score; \model{} is trained on the SciTS training split. \textbf{(b)} ST-Bench T4 forecasting, MAE; \model{} is trained on the T4 training split. \textbf{(c)} CiK forecasting, RCRPS; \model{} is evaluated zero-shot. Baseline results are from each benchmark.}
\label{fig:ood}
\vspace{-12pt}
\end{figure}

Figure~\ref{fig:ood} tests \model{} on three additional open benchmarks, each chosen for a different ability and scored under its native protocol. SciTS~\citep{scits} tests \model{}'s understanding performance beyond \dataset{}, across ten scientific domains. ST-Bench T4~\citep{streasoner} and CiK~\citep{cik} test \model{}'s context-aided forecasting performance beyond \dataset{}. On SciTS and ST-Bench T4, \model{} is first trained on their training splits. CiK has no training split, so \model{} as trained on \dataset{} is evaluated directly, which tests its generalization to unseen datasets. On all three, the \model{} architecture outperforms every dedicated time series model. On SciTS, \model{} leads every baseline on all ten domains, including TimeOmni trained on the same split (Figure~\ref{fig:ood}a, Table~\ref{tab:app-scits}), which shows that the architecture's advantage on understanding carries over beyond \dataset{}. On ST-Bench T4, \model{} has a lower MAE than all 14 rows the benchmark reports, including its own STReasoner-8B and the best zero-shot row (Figure~\ref{fig:ood}b). On CiK, \model{} outperforms every TSFM, multimodal forecaster, and statistical method the benchmark reports (Figure~\ref{fig:ood}c). Prompted LLMs far larger than \model{} (Llama-3-70B and GPT-4o) can be ahead of it, possibly because CiK's histories are short enough to be read as text, so their stronger text understanding decides the forecast. Prompted LLMs of similar or smaller size, such as Llama-3-8B, usually trail \model{}, as reading even these short histories as text may be difficult at that model size.

\section{Conclusion}
\label{sec:conclusion}

We presented \model{}, which bidirectionally couples a pretrained TSFM and an LLM for multimodal time series understanding and forecasting, so that one model architecture reads and generates both series and text, reasons in text, and draws on the TSFM's pretrained forecasting ability. We also built \dataset{}, a multimodal time series reasoning dataset with annotated reasoning traces that covers both understanding and forecasting tasks across 12 domains. The results support three findings. First, on \dataset{}, neither component alone suffices: The LLM struggles with long time series and forecasts poorly even after fine-tuning, and the TSFM reads no text, so it attempts only the closed-label understanding subtasks and forecasts without the event text. Second, coupling the two narrows both gaps: \model{} leads its components on understanding and forecasting tasks, and outperforms dedicated time series models on open benchmarks. Third, reasoning helps most when the output cannot be read off the input directly, as in counting events, discriminating among many anomaly classes, or deriving a forecast's direction and shape from the event.

\subsection*{AI use statement}
Generative AI is used in this work in three roles. First, the reasoning traces and forecast instructions of the forecasting task of \dataset{} are annotated by an LLM (Appendix~\ref{app:dataset}). Second, an LLM judge scores the generated reasoning (Section~\ref{sec:results-trace}, Appendix~\ref{app:trace-eval}). Third, LLM assistants are used to polish prose and to draft utility code. We review all AI-assisted work: the annotated traces pass the human review of Appendix~\ref{app:audit}, the judge is checked against a chance reference and a counterfactual edit, and all assisted text and code are reviewed by the authors, with the code covered by the released test suite.

\subsection*{Ethics statement}
\dataset{} is built entirely from publicly released datasets, used under their licenses; the physiological recordings among them are de-identified public corpora, and no data was newly collected from human subjects. The human quality review of Appendix~\ref{app:audit} was carried out by two doctoral students, with no personal data involved. The annotated reasoning traces are LLM-generated and can carry errors of the annotating model; their quality is reviewed in Appendix~\ref{app:audit}. The model produces forecasts and generated reasoning for research evaluation; deployment in an operational setting would require domain-specific validation.

\subsection*{Reproducibility statement}
The architecture is specified in Section~\ref{sec:model}, with the sliding-window Q-Former's window-planning rule and encodings in Appendix~\ref{app:swqformer}. The annotation of the forecasting task of \dataset{} is documented in Appendix~\ref{app:dataset}: the sources, the annotation protocol and its constraints. The human review protocol and its inter-reviewer agreement are reported in Appendix~\ref{app:audit}. All training scripts, evaluation pipelines, and the CPU test suite are part of the code base. The code is released at \url{https://github.com/OpenTSLab/TimeOmni-o1}. A project page presents dataset samples and case studies: \url{https://windowso.github.io/TimeOmni-o1-web/}.

\bibliography{chronos_llm}
\bibliographystyle{iclr2027_conference}

\appendix
\section{Appendix}

\subsection{Limitations}
\label{app:limitations}

The forecasting task of \dataset{} is much smaller than the understanding task, with 5{,}067 samples over five domains against 568{,}586. The forecasting traces of \dataset{} are annotated by an LLM, and human review covers 280 of the 5{,}067 samples (Appendix~\ref{app:audit}). The generated reasoning is scored by an LLM judge, which is checked against a chance reference but not against human raters.

The soft prompt approximates the numeric values of the time series, so on questions that ask for exact values of a short time series the fine-tuned LLM reading full-precision digits is ahead (Section~\ref{sec:results-main}). On CiK, prompted LLMs far larger than \model{} can be ahead (Section~\ref{sec:results-ood}).

\subsection{Positioning against related models and datasets}
\label{app:related-tables}

Tables~\ref{tab:app-model-cmp} and~\ref{tab:app-dataset-cmp} list the models and datasets discussed in Section~\ref{sec:related}, one yes-or-no property per column. Every cell follows the cited paper's own description of its method or dataset.

\textbf{Models.} Table~\ref{tab:app-model-cmp} lists five properties of each model. The models that forecast with a pretrained TSFM serve one task type: DoubleCast, ChronoSteer and STRIDE forecast but answer no questions, and TS-Reasoner answers questions but does not forecast. The seven TSLMs that serve both task types generate the forecast on the LLM side: as text, as tokens, through a prediction head or, in TimeOmni-VL, as a plot. None of them uses a pretrained TSFM. Three other models pass their reasoning to the forecaster. ChronoSteer maps the LLM's instruction to one of nine fixed instructions, STRIDE averages the LLM's hidden states into one vector, and TimeOmni-VL conditions an image generator whose output plot is converted back to values. \model{} passes the compressed LLM hidden states to every TSFM block.

\begin{table}[h]
\centering
\footnotesize
\setlength{\tabcolsep}{5pt}
\caption{\model{} against related time series models. \textbf{Und.}~/~\textbf{Fc.}: the model answers questions about a time series in text / forecasts. \textbf{Reasoning}: it generates a reasoning trace before its answer or forecast. \textbf{TSFM}: a pretrained TSFM is part of the model; \emph{optional}: one variant of OpenTSLM can use a frozen Chronos-2 in place of its own encoder. \textbf{Reasoning to forecaster}: the generated reasoning is passed to a separate forecaster.}
\label{tab:app-model-cmp}
\begin{tabular}{lccccc}
\toprule
Model & Und. & Fc. & Reasoning & TSFM & Reasoning to forecaster \\
\midrule
\multicolumn{6}{l}{\emph{Time series models without an LLM}} \\
Chronos-2 \citep{chronos2} & \xmark & \cmark & \xmark & \cmark & \xmark \\
UniTS \citep{units} & \xmark & \cmark & \xmark & \xmark & \xmark \\
\midrule
\multicolumn{6}{l}{\emph{Forecasters that couple an LLM with a pretrained TSFM}} \\
DoubleCast \citep{doublecast} & \xmark & \cmark & \xmark & \cmark & \xmark \\
ChronoSteer \citep{chronosteer} & \xmark & \cmark & \cmark & \cmark & \cmark \\
STRIDE \citep{stride} & \xmark & \cmark & \cmark & \cmark & \cmark \\
\midrule
\multicolumn{6}{l}{\emph{Time series language models (TSLMs)}} \\
ChatTime \citep{chattime} & \cmark & \cmark & \xmark & \xmark & \xmark \\
Chat-TS \citep{chattsllm} & \cmark & \xmark & \xmark & \xmark & \xmark \\
ChatTS \citep{chatts} & \cmark & \xmark & \xmark & \xmark & \xmark \\
Time-MQA \citep{timemqa} & \cmark & \cmark & \cmark & \xmark & \xmark \\
TimeReasoner \citep{timereasoner} & \xmark & \cmark & \cmark & \xmark & \xmark \\
Time-R1 \citep{timer1} & \xmark & \cmark & \cmark & \xmark & \xmark \\
TimeOmni-1 \citep{timeomni1} & \cmark & \cmark & \cmark & \xmark & \xmark \\
OpenTSLM \citep{opentslm} & \cmark & \xmark & \cmark & optional & \xmark \\
TsLLM \citep{tsllm} & \cmark & \cmark & \xmark & \xmark & \xmark \\
TimeOmni \citep{scits} & \cmark & \cmark & \xmark & \xmark & \xmark \\
STReasoner \citep{streasoner} & \cmark & \cmark & \cmark & \xmark & \xmark \\
TS-Reasoner \citep{tsreasoner} & \cmark & \xmark & \cmark & \cmark & \xmark \\
TimeOmni-VL \citep{timeomnivl} & \cmark & \cmark & \cmark & \xmark & \cmark \\
\midrule
\textbf{\model{} (Ours)} & \cmark & \cmark & \cmark & \cmark & \cmark \\
\bottomrule
\end{tabular}
\end{table}

\textbf{Datasets.} Table~\ref{tab:app-dataset-cmp} is the full version of Table~\ref{tab:dataset-cmp}: it adds the remaining source corpora of \dataset{} and keeps the same columns. Annotated traces exist in the six understanding sources and in TSR-Suite. Among these seven, two also include a forecasting task: ST-Bench, whose time series are synthetic, and TSR-Suite, where 2.3k of the 23.6k samples carry a trace. The forecasting sources pair time series with text but carry no traces. \dataset{} annotates every sample of both task types. It attains its size by unifying the six understanding sources under one schema and by annotating the ten forecasting subsets collected from the five forecasting sources.

\begin{table}[h]
\centering
\footnotesize
\setlength{\tabcolsep}{3pt}
\caption{\dataset{} against its source corpora and related multimodal time series datasets; columns as in Table~\ref{tab:dataset-cmp}. \emph{partial}: only a subset of the samples carries a trace (TelecomTS: the anomaly and network-level questions; TSR-Suite: 2.3k of 23.6k samples). --: the source reports no domain count, or releases continuous series rather than fixed samples.}
\label{tab:app-dataset-cmp}
\begin{tabular}{lccclccr}
\toprule
Dataset & Und. & Fc. & Trace & Domains & Series & Train split & Samples \\
\midrule
\multicolumn{8}{l}{\emph{Understanding sources of \dataset{}}} \\
OpenTSLM CoT sets \citep{opentslm} & \cmark & \xmark & \cmark & 1 & real & \cmark & 326k \\
ST-Bench \citep{streasoner} & \cmark & \cmark & \cmark & 10 & synthetic & \cmark & 28.2k \\
RATs40K \citep{timera} & \cmark & \xmark & \cmark & 10 & real & \cmark & 39.6k \\
TSRBench \citep{veritime} & \cmark & \xmark & \cmark & 4 & both & \cmark & 4.3k \\
TelecomTS \citep{telecomts} & \cmark & \xmark & partial & 1 & real & \cmark & 2.2M \\
HiTSR \citep{hitsr} & \cmark & \xmark & \cmark & -- & both & \cmark & 83.8k \\
\midrule
\multicolumn{8}{l}{\emph{Forecasting sources of \dataset{}}} \\
CGTSF \citep{chattime} & \xmark & \cmark & \xmark & 3 & real & \cmark & 33.7k \\
Fidel-TS \citep{fidelts} & \xmark & \cmark & \xmark & -- & real & \cmark & -- \\
MTBench \citep{mtbench} & \cmark & \cmark & \xmark & 2 & real & \xmark & 22k \\
News Forecast \citep{newsforecast} & \xmark & \cmark & \xmark & 4 & real & \cmark & -- \\
Time-MMD \citep{timemmd} & \xmark & \cmark & \xmark & 9 & real & \cmark & 17.1k \\
\midrule
\multicolumn{8}{l}{\emph{Other related datasets}} \\
TSQA \citep{timemqa} & \cmark & \cmark & \xmark & 12 & both & \cmark & 193k \\
TSR-Suite \citep{timeomni1} & \cmark & \cmark & partial & 10 & real & \cmark & 23.6k \\
CiK \citep{cik} & \xmark & \cmark & \xmark & 7 & both & \xmark & 355 \\
SciTS \citep{scits} & \cmark & \cmark & \xmark & 12 & both & \cmark & 52.1k \\
WIT \citep{whatiftsf} & \xmark & \cmark & \xmark & 4 & real & \xmark & 5.4k \\
\midrule
\textbf{\dataset{} (Ours)} & \cmark & \cmark & \cmark & 12 & both & \cmark & 574k \\
\bottomrule
\end{tabular}
\end{table}

\subsection{Sliding-window Q-Former: planning rule and encodings}
\label{app:swqformer}

\textbf{Window planning and local compression.} The window count is set before the patch sequence is partitioned, as $W=\min\bigl(W_{\min}+b\log_2 N,\,W_{\max}\bigr)$ with $N=P\times C$, rounded to the nearest integer and capped at $P$ so that every window spans at least one patch position. Among these quantities, $k$, $W_{\min}$, $W_{\max}$ and the number of global queries $Q_{\mathrm{g}}$ are hyperparameters, and $b$ is set by the $N$ at which $W$ reaches $W_{\max}$. Window $i$ spans the patch positions $[\,i\lceil P/W\rceil,\,(i{+}1)\lceil P/W\rceil)$ of every channel, so it holds $C\lceil P/W\rceil$ patch representations, which equals the window width $\lceil N/W\rceil$ of Section~\ref{sec:series2llm} whenever $W$ divides $P$; the final window is zero-padded and masked. Each window is summarized by its $k$ queries through a Q-Former shared by every window, with the window's patch representations as keys and values. The $W$ summaries are concatenated in time order, so the order of the windows is given by their position in the output sequence, and the order inside a window is given by the position embedding below. The $Q_{\mathrm{g}}$ global queries attend to all $N$ patches in one cross-attention and are prepended to the local tokens, so the Q-Former outputs $Q_{\mathrm{g}}+kW$ tokens.

\textbf{Encodings.} The learned channel embedding is added to each channel's patches before the partition into windows, so the global branch reads the same channel-embedded patches as the local one. The window-relative position embedding is indexed by the offset $0,\dots,\lceil P/W\rceil-1$ and restarts in every window, so it encodes order inside a window only; its table is capped in length, and any longer window reuses the last row. The global branch spans the whole history, which a capped learned table cannot cover, so its keys instead carry a parameter-free sinusoidal time encoding, evaluated at each patch's real position in the original time series. The channels of one sample are aligned in time, so patches at the same position share this encoding and differ only by their channel embedding. Normalization uses one $(\mu,\sigma)$ pair per channel, taken over its whole history, so all windows of a channel share one scale.

\subsection{Annotation of the forecasting task of \dataset{}}
\label{app:dataset}

\textbf{Sources.} Table~\ref{tab:app-fc-provenance} lists the ten subsets, their sources, their sampling resolutions and their window lengths.

\begin{table}[h]
\centering
\footnotesize
\caption{Provenance of the forecasting task of \dataset{}. Time series are taken from ten subsets of five public multimodal forecasting sources; the background, event description, reasoning trace, forecast instruction and ROI are annotated in this work. History and Horizon: window lengths in steps; the equities windows vary with the trading session and are reported as medians. Sources as cited in Section~\ref{sec:dataset-fc}.}
\label{tab:app-fc-provenance}
\setlength{\tabcolsep}{2pt}
\begin{tabular}{llllrr}
\toprule
Source & Subset & Domain & Resolution & History & Horizon \\
\midrule
CGTSF & MSPG, Melbourne PV & Solar & 15 min & 480 & 96 \\
CGTSF & PTF, Paris traffic & Traffic & hourly & 120 & 24 \\
Fidel-TS & 50Hertz renewable & Solar & 15 min & 288 & 96 \\
Fidel-TS & Calgary PV & Solar & hourly & 96 & 24 \\
Fidel-TS & California ISO & Load & 5 min & 576 & 288 \\
MTBench & US equities & Finance & intraday & 390 & 78 \\
News Forecast & Australian load & Load & 30 min & 48 & 48 \\
News Forecast & California traffic & Traffic & hourly & 120 & 48 \\
News Forecast & Bitcoin & Finance & daily & 60 & 14 \\
Time-MMD & US drought & Climate & weekly & 36 & 12 \\
\bottomrule
\end{tabular}
\end{table}

\textbf{Annotation protocol.} The 5{,}067 collected samples are annotated by an LLM annotator, Claude Sonnet 4.6.\footnote{\url{https://www.anthropic.com/news/claude-sonnet-4-6}} Annotation runs in two stages, separated by what each field is used for. The event description is a model input at test time, so the first stage must not depend on the future. In this stage the annotator receives a hand-written dataset background, the sample's accompanying text and the timestamps of the history and the horizon; it receives neither the time series nor any future value. The accompanying text is the source's own aligned text rather than retrieved material: weather forecasts and reports for the solar and load sets, aligned news for the finance and traffic sets, and drought reports for the climate set. Text dated inside the forecast horizon is marked, and the annotator is instructed to ignore it. The annotator returns the event description, condensed from the accompanying text without causal phrasing. The reasoning trace, the forecast instruction and the ROI are training targets and are written by the annotator in the second stage. The instruction names the ROI and states the direction, shape and relative magnitude of the effect in words. Here the annotator receives the background, the event description of the first stage as a fixed input, the window metadata, a rendering of the time series, and the future window as a timestamped value table. Three constraints are imposed on the second stage: the ROI lies inside the horizon, the trace uses the history and the event only, and no future value is cited. A schema check enforces the ROI bounds, and an automatic check flags any quoted future value.

\textbf{Annotation prompts.} Prompts~\ref{box:fc-event} and~\ref{box:fc-trace} reproduce the annotator's prompts for the two stages, each as a system message followed by the user message template. Fields in braces, such as \texttt{\{background\}}, are filled per sample, and the output schema is shown at the position it occupies in the annotator's input. The \texttt{event} field of the first stage holds the event description and is passed unchanged to the second stage. There, the \texttt{reasoning} and \texttt{conclusion} fields hold the reasoning trace and the forecast instruction; \texttt{answer} carries the ROI bounds and a shape phrase.

\begin{promptbox}[label={box:fc-event}]{Stage 1 of the forecasting-task annotator: event description}
[System message]

You are a synthetic-data annotator for time-series forecasting research.

You are shown a pre-written dataset background, the real ACCOMPANYING TEXT for
one sample (a weather forecast / industry report / news / covariate
description), and the timestamps of the history window and of the forecast
horizon. You are NOT shown the time series, and you have no information about
what happens in the forecast horizon. The accompanying text is a REAL
event/context that is available at forecast time — you do NOT invent it.
Produce ONE JSON object (schema given in the user message).

SUMMARIZE the accompanying text into the `event` field: at most 500 words,
PURELY OBJECTIVE (only WHAT happened / WHEN / WHERE / WHO-WHAT). Keep the
statements that concern the measured quantity, its location, or the dates of
the forecast horizon (forecast conditions, scheduled or announced events,
calendar facts), and drop boilerplate/noise. Use ONLY forecast-time-available
content: if the text marks some lines `IN_FUTURE_WINDOW` / `LEAKAGE`, you MUST
ignore those lines (they are not knowable at forecast time). Do NOT add facts
the text does not state, do NOT predict the series, and do NOT write
impact/causal phrasing ("the household hosted a gathering", NOT "...which
raised demand"; avoid "causing", "leading to", "drove ... up"). If the text
carries no statement of this kind, say so plainly and briefly in `event`.

Global constraints:
- All free-text is plain ASCII/UTF-8, no markdown.
- Output strictly valid JSON only — no code fences, no prose before or after.

[User message template]

## Dataset background (given — do NOT rewrite)

{background}

## Accompanying real-world text (EVENT SOURCE — summarize into `event`, <=500 words)

This is the REAL text attached to this sample, available at forecast time. Lines
marked `IN_FUTURE_WINDOW` / `LEAKAGE` (if any) are NOT knowable at forecast time —
ignore them.

{text_body}

## Per-sample metadata

- timestamps: start={ts_start}, boundary={ts_boundary}, end={ts_end}
  (history window = start .. boundary; forecast horizon = boundary .. end)

## Output

Produce exactly this JSON (no extra/missing keys):

{
  "event":       "<=500-word OBJECTIVE summary of the accompanying text (WHO/WHAT/WHEN/WHERE); forecast-time info only; no impact/causal phrasing; no invented facts>"
}
\end{promptbox}

\begin{promptbox}[label={box:fc-trace}]{Stage 2 of the forecasting-task annotator: ROI, forecast instruction and reasoning trace}
[System message]

You are a synthetic-data annotator for explainable time-series forecasting research.

You are shown one time-series chart, a pre-written dataset background, the EVENT
description of this sample (written beforehand from the sample's real
accompanying text, available at forecast time), per-sample metadata, and a
`timestamp | idx | value` table of the FUTURE window. The event is GIVEN: do NOT
rewrite it, extend it, or add facts to it. Produce ONE JSON object (schema
given in the user message) by following these steps:

1. Locate ONE region of interest (ROI): a contiguous span lying ENTIRELY inside
   the future window (it may not start in the past window or straddle the
   forecast horizon). Pick the exact `start_idx` / `end_idx` from the
   future-window table. The span is HALF-OPEN `[start_idx, end_idx)`: it covers
   rows `start_idx` through `end_idx - 1` INCLUSIVE; the `end_idx` row is OUTSIDE
   the ROI. Read every clock time and shape feature ONLY from rows `start_idx` ..
   `end_idx - 1`. If a movement you want inside the ROI sits at row `end_idx`,
   extend `end_idx` so that point is included. Choose the span the event most
   plausibly governs; keep it as TIGHT as the effect — do NOT default to the
   whole future window.
2. State a CONCLUSION (one or two sentences) naming the index range
   `[start_idx, end_idx)` AND the DIRECTION, SHAPE and RELATIVE MAGNITUDE of the
   effect inside it. State the relative magnitude in words, as a comparison with
   the levels of the HISTORY window (e.g. "well below the series' typical highs",
   "close to its usual peak level"). Do NOT quote raw numbers (no "reaches 0.88",
   no "around 1.5 kWh").
3. Emit the machine-checkable ANSWER: `start_idx` / `end_idx` (the span from step
   1) and a `shape` phrase that faithfully and SPECIFICALLY describes the shape
   effect (do not genericize). `shape` is about the shape only — never the
   cause/event, never absolute numbers. Use "crest"/"peak"/"trough" only if the
   rise-then-fall (or fall-then-rise) reversal actually occurs WITHIN the ROI; if
   the values move monotonically to the last included row, say "ending at its
   highest/lowest point". If the ROI is essentially flat / low-level noise, say
   so plainly (e.g. "holds roughly level with minor fluctuation") and pick a
   minimal, neutral shape — do NOT manufacture a crest/surge/step out of baseline
   jitter; a calm, near-flat span is a valid annotation.
4. LAST, write the REASONING: a concise FORWARD chain (2-4 short sentences),
   treating the event as an ESTABLISHED FACT, with two links: (i) give the ROI's
   clock span as the timestamp at row `start_idx` through the timestamp at row
   `end_idx - 1` (the LAST included row) — NEVER the `end_idx` row's time — so
   the event's time lines up with it; and (ii) derive the direction, shape and
   relative magnitude from the event + the HISTORY window (`[0, boundary_idx)`),
   mapping event stages to shape sub-features when both have stages. Go beyond
   restating the event; do NOT hedge ("plausible", "would fit"), do NOT cite any
   future value, and do NOT add new facts beyond the event or history.

Global constraints:
- NO VISUAL VOCABULARY in any text field (conclusion / answer.shape /
  reasoning): the downstream consumer is text-only and never sees the chart.
  Forbidden words include "blue", "red", "color", "line", "curve", "plot",
  "chart", "figure", "image", "axis", "tick", "dashed", "x-axis", "y-axis".
  Refer to data only as "history window" / "future window" or by index range.
- Do NOT quote any `value` from the future-window table anywhere; relative
  magnitude is stated in words only.
- All free-text is plain ASCII/UTF-8, no markdown.
- Output strictly valid JSON only — no code fences, no prose before or after.

[User message template]

## Dataset background (given — do NOT rewrite)

{background}

## Event description (given — do NOT rewrite)

{event}

## Per-sample metadata

- past_len / future_len / total_len: {past_len} / {future_len} / {total_len}
- history window = indices [0, {past_end}); future window = [{past_end}, {total_len})
- ROI must satisfy {past_end} <= start_idx < end_idx <= {total_len}
  (half-open: the ROI covers rows start_idx .. end_idx-1; the end_idx row is excluded)
- timestamps: start={ts_start}, boundary={ts_boundary}, end={ts_end}

## Chart

Plots the series over time: history ({past_len} pts) then future ({future_len}
pts), split by the forecast-horizon line. The top axis carries the integer index
`i` (0..{last_idx}) of the concatenated `(past ++ future)` sequence — use it for
ROI indices. The image is for YOUR perception only.

## Future window values ({future_len} rows, `timestamp | idx | value`)

Pick the exact ROI from this table — `idx` is authoritative (do not shift it to
fit a time calc), `timestamp` gives the event's WHEN, `value` is for your read
of direction, shape and relative magnitude only (never quote it). The ROI's end clock time is the `timestamp` of row
`end_idx - 1` (the last included row), NOT the `end_idx` row.

{future_table}

## Output

Follow the system procedure and produce exactly this JSON (no extra/missing keys):

{
  "conclusion":  "<1-2 sentences: the index range [start_idx, end_idx) + the direction, shape and relative magnitude of the effect; no numbers>",
  "answer": {
    "start_idx":      <int>,
    "end_idx":        <int>,
    "shape":          "<faithful, specific phrase for the shape effect; no numbers, no visual words>"
  },
  "reasoning":   "<2-4 sentences, forward chain: (i) ROI clock time lines up with the event; (ii) event + history -> direction, shape, relative magnitude>"
}
\end{promptbox}
\subsection{Samples from \dataset{}}
\label{app:samples}

This section shows one sample per domain of the understanding task (U01 to U12) and one per domain of the forecasting task (F01 to F05). Questions and traces are abridged: option lists, category inventories and formatting instructions are shortened, and the values cited by the traces are checked against the stored series. Each understanding card shows the time series, the question, the annotated reasoning trace and the answer. Each forecasting card shows the history and the future ground truth with the ROI shaded, then the dataset background, the event description, the reasoning trace and the forecast instruction. Markers in the figures mark the values the trace cites.

\begin{samplebox}{Samples from \dataset{}: understanding U01 to U12, forecasting F01 to F05}

\tcbsubtitle{U01 \enspace Physiology \enspace (OpenTSLM ECG-QA)}\nopagebreak
\emph{Open-ended diagnostic question answering; 12-lead ECG, 10 s at 500 Hz (5,000 points per lead).}\par\nopagebreak\smallskip
\noindent\includegraphics[width=\linewidth]{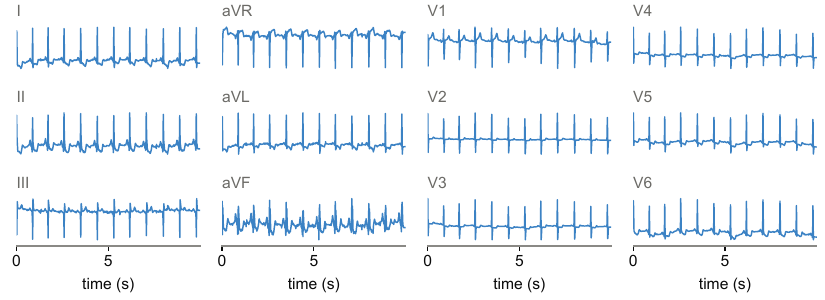}\par\smallskip
\textbf{Question.} Clinical context: 72-year-old female; 12-lead clinical ECG (baseline drift, static/burst noise and electrode artefacts noted; extra beats; pacemaker present). Which diagnostic symptom does this ECG show: myocardial infarction in inferoposterolateral leads, or myocardial infarction in anteroseptal leads (including uncertain symptoms)?\par\smallskip
\textbf{Reasoning trace.} Baseline drift and noise may obscure details and the pacemaker can affect some features, but the focus is on infarction patterns. The notable changes appear in the inferoposterolateral leads (II, III, aVF and possibly V5-V6), which show ST-segment elevation or other ischaemic changes; an anteroseptal infarction would instead involve V1-V4. Together with the clinical context this supports an inferoposterolateral myocardial infarction.\par\smallskip
\textbf{Answer.} myocardial infarction in inferoposterolateral leads

\tcbsubtitle{U02 \enspace Telecommunications \enspace (TelecomTS)}\nopagebreak
\emph{Network-level activity recognition; 16 KPI channels over 128 steps.}\par\nopagebreak\smallskip
\noindent\includegraphics[width=\linewidth]{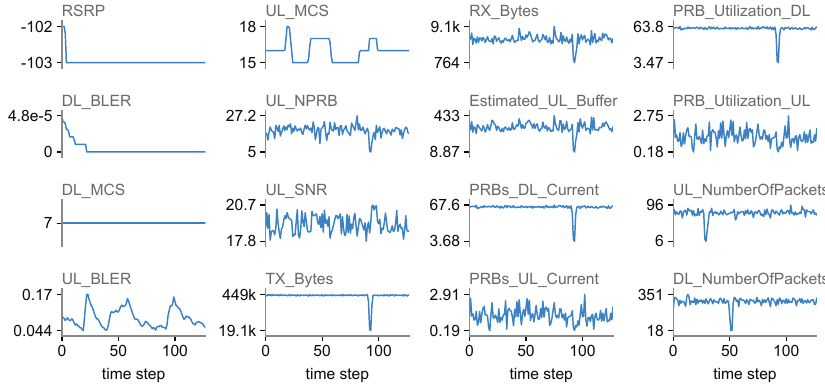}\par\smallskip
\textbf{Question.} A 5G monitoring window with 16 KPI channels. Observation summary (abridged): RSRP stable but weak ($-$103 to $-$102 dB); uplink SNR 17.8-20.7 dB; UL\_MCS 15-18 while DL\_MCS is constant at 7; DL\_BLER negligible, UL\_BLER 0.044-0.17; downlink PRBs about 63 with a sharp drop to single digits late in the trace, uplink PRB use minimal; traffic uplink-dominant and bursty (TX\_Bytes 19 k-449 k with a sharp dip, RX\_Bytes 0.8 k-9 k); the estimated uplink buffer is often large (mean 296 KB); 100\% TCP in both directions. Which type of activity occurred: YouTube viewing, Twitch stream, or file transfer?\par\smallskip
\textbf{Reasoning trace.} The radio link is stable but weak (RSRP about $-$103 dB) with a consistent uplink SNR (about 19.2 dB); uplink MCS varies slightly while downlink MCS stays constant. Traffic is highly uplink-bursty, with TX\_Bytes dropping sharply, and the estimated uplink buffer stays high, indicating queued data. This pattern is consistent with a file transfer.\par\smallskip
\textbf{Answer.} File download was the activity.

\tcbsubtitle{U03 \enspace Synthetic \enspace (HiTSR, level 3)}\nopagebreak
\emph{Multiple choice matching pattern, values and semantics; univariate, 480 monthly points.}\par\nopagebreak\smallskip
\noindent\includegraphics[width=\linewidth]{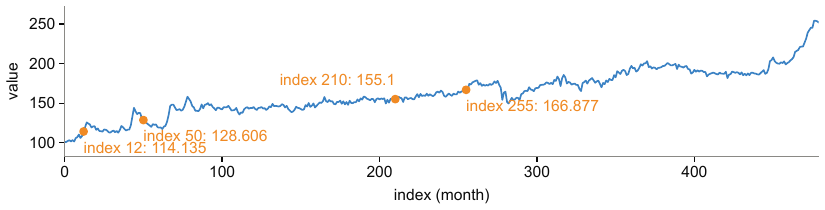}\par\smallskip
\textbf{Question.} Which statement best describes the long-term behaviour of this monthly retail broiler composite series and confirms a specific value? A. A clear upward trend with no seasonal component; the value at index 255 is 166.877. B. Strong periodic seasonality with a 12-month cycle; the value at index 12 is 114.135. C. A monotonic decrease after index 200; the value at index 210 is 155.1. D. Roughly constant around 150 for the first 100 months; the value at index 50 is 128.606.\par\smallskip
\textbf{Reasoning trace.} A: the series rises over the long run, most steeply beyond index 250, with no consistent seasonal pattern; index 255 is exactly 166.877; a gradual price increase in the broiler market is plausible. B: peaks and troughs do not recur annually, so the 12-month seasonality claim fails even though 114.135 at index 12 is correct. C: values climb above 250 late in the series, contradicting a monotonic decline, although 155.1 at index 210 is correct. D: the series starts near 100 and drifts up to about 140 by index 100, so it is not constant around 150; 128.606 at index 50 is correct.\par\smallskip
\textbf{Answer.} A

\tcbsubtitle{U04 \enspace Urbanism \enspace (ST-Bench)}\nopagebreak
\emph{Spatio-temporal graph reasoning, entity identification; 5 graph nodes over 48 steps with edges 0$\to$2, 1$\to$2, 2$\to$3, 3$\to$4, 4$\to$2.}\par\nopagebreak\smallskip
\noindent\includegraphics[width=\linewidth]{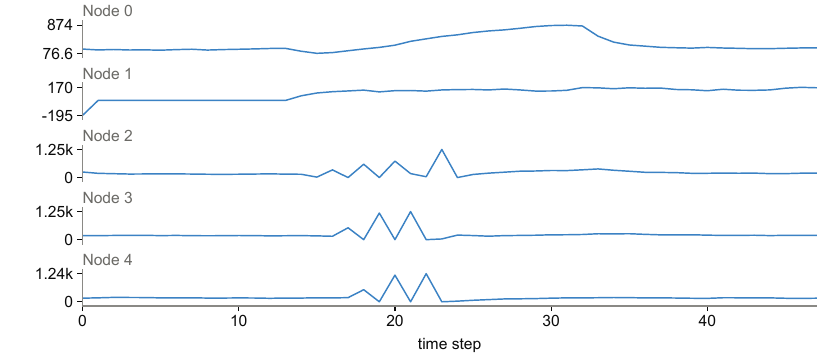}\par\smallskip
\textbf{Question.} Five nodes each carry a length-48 series; graph edges: Node 0$\to$2, 1$\to$2, 2$\to$3, 3$\to$4, 4$\to$2. Which (name, description) pair should Node 3 correspond to? A. Central Square, main city square intersection; B. Highway Exit, major highway exit point; C. Ring Road, ring road junction; D. Bypass Junction, peripheral traffic bypass point.\par\smallskip
\textbf{Reasoning trace.} Graph: Node 2 is a convergence point fed by Nodes 0 and 1, then flows to Node 3 $\to$ Node 4 $\to$ back to Node 2 (a loop). Series: Node 0 surges from about 76 to about 874 around indices 15-33; Node 1 is zero for indices 1-13 and then holds 60-170; Node 2 is erratic with extreme spikes (1250, 731, 593); Node 3 has a stable 150-200 baseline, extreme spikes (1184, 1250) near indices 17-23, then stabilises higher (180-265); Node 4 mirrors Node 3 with spikes (1242, 1178) followed by very low values (28, 66, 96). Node 3 receives the converged flow, carries the major surge, and sits on the circular 2$\to$3$\to$4$\to$2 structure, so it is a junction on a ring road rather than the central convergence point itself.\par\smallskip
\textbf{Answer.} C (Ring Road, ring road junction)

\tcbsubtitle{U05 \enspace Information technology \enspace (TSRBench)}\nopagebreak
\emph{Rule-based anomaly detection; univariate, 256 points (Kubernetes cluster autoscaler metric).}\par\nopagebreak\smallskip
\noindent\includegraphics[width=\linewidth]{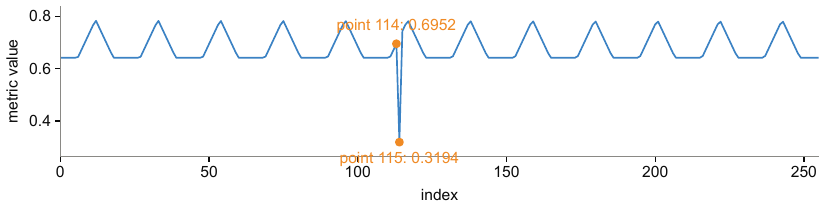}\par\smallskip
\textbf{Question.} Cluster Autoscaler Metrics collected from a Kubernetes cluster, length 256. Given that a sudden drop of more than 0.3 is considered anomalous, should the series be considered anomalous near point 114?\par\smallskip
\textbf{Reasoning trace.} Step 1: this is anomaly detection with an explicit rule (a drop larger than 0.3). Step 2: the relevant patterns are the drop magnitude, the 0.3 threshold and the location near point 114. Step 3: the value at index 114 is 0.6952 and the next value (index 115) is 0.3194, a drop of 0.3758 that exceeds the threshold; the drop is abrupt relative to the cyclical pattern (peaks about 0.78, troughs about 0.64). Steps 4-5: the gradual cyclical rises and falls do not explain such an abrupt drop, so the rule is unambiguously violated. Step 6: a drop of 0.3758 directly adjacent to point 114 qualifies as an anomaly.\par\smallskip
\textbf{Answer.} Yes

\tcbsubtitle{U06 \enspace Neuroscience \enspace (OpenTSLM Sleep)}\nopagebreak
\emph{Sleep-stage classification; single-channel EEG, one 30-s epoch (1,500 points).}\par\nopagebreak\smallskip
\noindent\includegraphics[width=\linewidth]{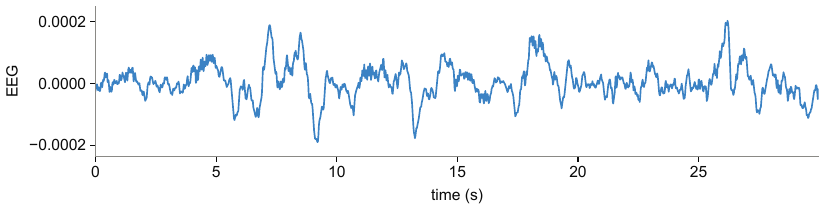}\par\smallskip
\textbf{Question.} A 30-second EEG segment corresponds to one of two sleep stages: Non-REM stage 4 or Wake. Determine the stage from the signal alone, reasoning in a single paragraph and ending with the answer.\par\smallskip
\textbf{Reasoning trace.} The signal shows relatively slow oscillations with large amplitude variations, indicating synchronised neuronal activity. Such slow-wave activity is the hallmark of deep sleep; a wakeful state would instead show faster, lower-amplitude, desynchronised activity. The segment therefore matches a deep sleep stage rather than wakefulness.\par\smallskip
\textbf{Answer.} Non-REM stage 4

\tcbsubtitle{U07 \enspace Earth science \enspace (RATs40K)}\nopagebreak
\emph{21-way anomaly diagnosis; univariate, 64 points (environmental sensor).}\par\nopagebreak\smallskip
\noindent\includegraphics[width=\linewidth]{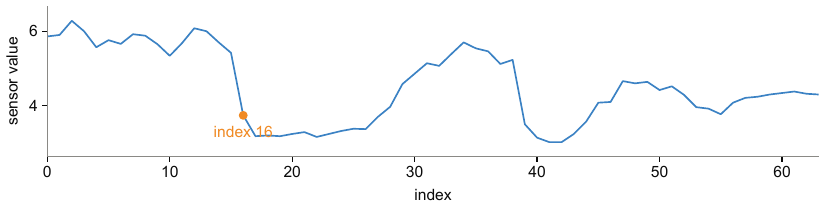}\par\smallskip
\textbf{Question.} A univariate series of length 64 collected from an environmental sensor. Analyse the temporal dynamics and name the category that best describes it, choosing from: Amplitude, Change Point, Collinearity Shift, Continuous Segment, Covariance Structure, Distributional Change, Drift, Joint Space, Nonlinear Pattern, Pattern Change, Periodic Change, Point, Principal Component Space, Repeated Value, Sparse, Sudden Flatline, Sudden Spike, Temporal Dependency, Trend Change, Trend Divergence Anomaly, or Normal Sequence.\par\smallskip
\textbf{Reasoning trace.} The series initially fluctuates within a stable range (about 5.5-6.3). Starting at index 16 the values drop to a lower range (about 3.0-3.4), an abrupt shift in both mean and variance that marks a change point. There is some later recovery and fluctuation, but the initial drop is a clear change in statistical behaviour.\par\smallskip
\textbf{Answer.} Change Point Anomaly

\tcbsubtitle{U08 \enspace Meteorology \enspace (RATs40K)}\nopagebreak
\emph{Multivariate anomaly diagnosis, 7-way; 3 variables over 32 steps (Tropical Atmosphere Ocean buoy array).}\par\nopagebreak\smallskip
\noindent\includegraphics[width=\linewidth]{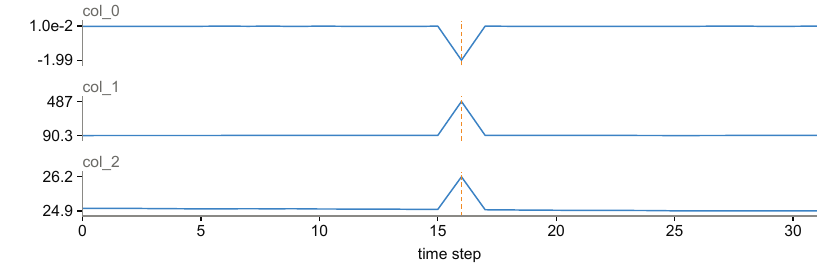}\par\smallskip
\textbf{Question.} A multivariate series with 3 variables (col\_0, col\_1, col\_2), each of length 32, collected from ocean / tropical-atmosphere buoy observations. Which category best describes it: Collinearity Shift, Covariance Structure, Joint Space, Principal Component Space, Temporal Dependency, or Trend Divergence Anomaly, or Normal Sequence?\par\smallskip
\textbf{Reasoning trace.} col\_0 stays near zero with small fluctuations but drops abruptly to $-$1.99 at time step 16. col\_1 progresses smoothly but spikes to 487.0 at the same step. col\_2 trends gradually downward but jumps to 26.2 at step 16. Each variable is individually anomalous at step 16, and their joint configuration there is inconsistent with the expected multivariate relationships, indicating a system-level disturbance.\par\smallskip
\textbf{Answer.} Joint Space Anomaly

\tcbsubtitle{U09 \enspace Economics \enspace (TSRBench)}\nopagebreak
\emph{Multi-step numerical calculation, counting events; univariate, 256 daily points.}\par\nopagebreak\smallskip
\noindent\includegraphics[width=\linewidth]{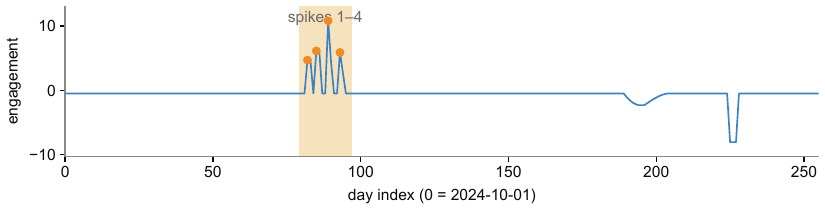}\par\smallskip
\textbf{Question.} Daily Customer Engagement (Marketing \& Sales), 256 points starting 1 October 2024. The company ran a series of marketing campaigns; during the period engagement showed a series of rapid spikes each followed by a quick decline. How many such spike-and-decline events occurred, and what could have caused them?\par\smallskip
\textbf{Reasoning trace.} Step 1: count distinct `rapid spike followed by quick decline' events. Step 2: a spike must clearly exceed the baseline (about $-$0.47), consecutive spike days form one event, and the decline must return to baseline within 1-2 days. Step 3: spike 1 on days 82-83 (4.71, 4.71, then $-$0.48); spike 2 on days 85-86 (6.13, 6.14, then $-$0.47); spike 3 on days 89-90 (10.82, 4.30, then $-$0.48); spike 4 on days 93-94 (5.90, 2.71, then $-$0.47). Step 4: four events. Step 5: no overlapping events; the later negative excursions (e.g. $-$8.05) are downward anomalies, not engagement spikes. Step 6: four spike-decline events, consistent with campaign launches.\par\smallskip
\textbf{Answer.} 4

\tcbsubtitle{U10 \enspace Bioacoustics \enspace (TSRBench)}\nopagebreak
\emph{Binary classification, call present or absent; 2-s hydrophone waveform.}\par\nopagebreak\smallskip
\noindent\includegraphics[width=\linewidth]{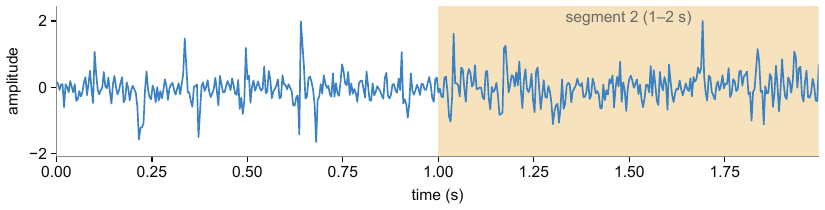}\par\smallskip
\textbf{Question.} An audio segment may contain a North Atlantic right whale up-call (about 60-250 Hz, lasting about 1 s); the low-frequency band is often congested with ship noise, drilling or piling. The 2-second waveform is sampled at 2 kHz. Does the segment contain a right whale call? (a) Right Whale Present, (b) No Right Whale.\par\smallskip
\textbf{Reasoning trace.} Step 1 (task intent): binary classification of whether a 2-s segment contains an up-call defined by its 60-250 Hz band and about 1 s duration. Step 2 (key patterns): frequency range, duration, amplitude consistency (a gradually rising `up-call' envelope rather than impulsive noise), and anthropogenic interference. Step 3 (analysis): split the waveform into two 1-s segments. Segment 1 (0-1 s) has scattered in-band energy and strong energy below 60 Hz, typical of ship noise. Segment 2 (1-2 s) shows a sustained energy concentration at 90-180 Hz for about 1 s, matching the band and duration of an up-call. Step 4: segment 2 satisfies the band and duration criteria. Step 5 (reflection): the largest excursions of segment 1 (near 0.2 s and 0.7 s) are short and broadband, not a sustained tone; segment 1 lacks spectral focus. Step 6: the sustained, band-limited, about-1-s component in the second half identifies an up-call.\par\smallskip
\textbf{Answer.} Right Whale Present

\tcbsubtitle{U11 \enspace Manufacturing \enspace (RATs40K)}\nopagebreak
\emph{21-way anomaly diagnosis; univariate, 64 points (machine system).}\par\nopagebreak\smallskip
\noindent\includegraphics[width=\linewidth]{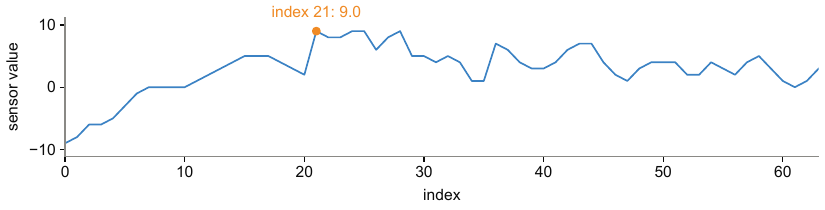}\par\smallskip
\textbf{Question.} A univariate series of length 64 collected from a machine system. Analyse the temporal dynamics and name the category that best describes it (same 21 categories as above).\par\smallskip
\textbf{Reasoning trace.} Several irregularities are present: a flatline of zeros at indices 7-10, repeated values (5.0 at indices 15-17, 8.0 at indices 22-23), and fluctuating segments without a consistent trend. The most prominent, however, is the sudden jump to 9.0 at index 21.\par\smallskip
\textbf{Answer.} Sudden Spike Anomaly

\tcbsubtitle{U12 \enspace Energy \enspace (TSRBench)}\nopagebreak
\emph{Appliance-signature classification; univariate, 720 points (2-min sampling over 24 h).}\par\nopagebreak\smallskip
\noindent\includegraphics[width=\linewidth]{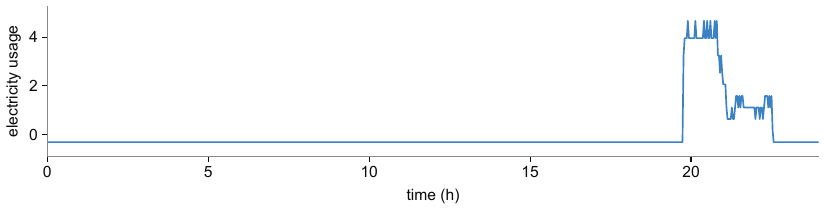}\par\smallskip
\textbf{Question.} Electricity usage of a UK household from the `Powering the Nation' study, sampled every 2 minutes over 24 hours (720 points). Classify the device usage pattern: (a) Desktop or (b) Laptop.\par\smallskip
\textbf{Reasoning trace.} Step 1: binary classification of a device energy signature. Step 2: key patterns are the baseline level, the duration of sustained high consumption, and volatility. Step 3: the signal sits at one repeated low value ($-$0.3151) for 636 of the 720 points; the only active block runs from index 593 to 676 (about 168 minutes), holds 3.96 to 4.67 for about 62 minutes, then steps down through 3.25, 2.06, 1.11 and 0.64; transitions such as 1.59 $\to$ $-$0.32 are abrupt. Step 4: an idle baseline for most of the day, a single active block of under three hours and stepped power levels match a laptop (battery-driven power management) rather than a desktop's steady higher load over a working day. Step 5: the 4.67 readings are single 2-minute points inside the block, not a sustained higher load. Step 6: Laptop.\par\smallskip
\textbf{Answer.} Laptop

\tcbsubtitle{F01 \enspace Solar \enspace (CGTSF, MSPG Melbourne PV)}\nopagebreak
\emph{Event-conditioned forecasting; 15-min resolution, history 480 and horizon 96 points; ROI [514, 552), 08:30 to 17:45 on 8 March 2021.}\par\nopagebreak\smallskip
\noindent\includegraphics[width=\linewidth]{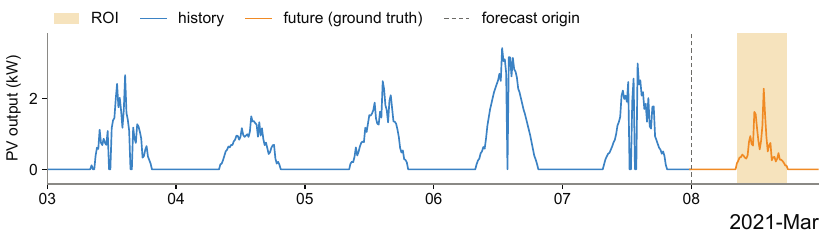}\par\smallskip
\textbf{Background.} Power generated by a single photovoltaic site in Melbourne, Australia, every 15 minutes. Output follows a daily bell shape (zero overnight, peak near solar noon) whose height and smoothness are gated by cloud, rain and haze.\par\smallskip
\textbf{Event.} Monday, 8 March 2021 is the target day. Conditions: moderate drizzle with a gentle breeze, 14-19 C; sunrise 06:11, sunset 18:50.\par\smallskip
\textbf{Reasoning trace.} The span from 08:30 (row 514) through 17:45 (row 551) covers the generation window of the target day, inside the sunrise-to-sunset interval given by the event. Moderate drizzle through the whole day places it among the low days of the history window, whose output stays well below the highest day. Those low days do not form a smooth bell: output is uneven all day, with isolated readings well above their neighbours. The target day follows the same profile, low and erratic with at most brief isolated higher readings, and falls to zero before sunset.\par\smallskip
\textbf{Forecast instruction.} Over index range [514, 552), solar generation remains at a persistently low, erratic level throughout the active generation window, with a brief pronounced peak in the early afternoon forming the single highest reading of the day, before declining back to low irregular values and reaching zero before sunset. Overall, with drizzle reported, the peak stays well below the series' typical highs.

\tcbsubtitle{F02 \enspace Load \enspace (Fidel-TS, California ISO)}\nopagebreak
\emph{Event-conditioned forecasting; 5-min resolution, history 576 and horizon 288 points; ROI [720, 781), 12:00 to 17:00 on 30 August 2019.}\par\nopagebreak\smallskip
\noindent\includegraphics[width=\linewidth]{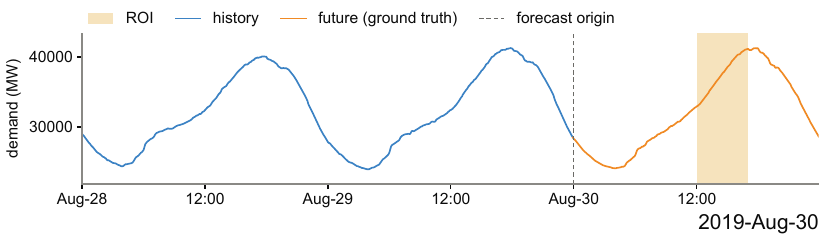}\par\smallskip
\textbf{Background.} California ISO grid electricity demand in megawatts, every 5 minutes. The dominant pattern is a daily cycle (overnight trough, daytime rise, evening peak) shaped by temperature-driven heating and cooling load.\par\smallskip
\textbf{Event.} Forecast for California on 30 August 2019: warm and sunny, 18-23 C, partly sunny with a gentle west breeze; temperatures rise through the day to an afternoon high, then ease slightly in the evening under clear skies.\par\smallskip
\textbf{Reasoning trace.} The selected span runs from 12:00 to 17:00 on 30 August, matching the afternoon period when temperatures keep rising to their daily high. In the history window demand follows a recurring daytime build toward a late-day high. With warm, sunny conditions persisting into the afternoon, the future window keeps climbing through this interval and finishes at its highest point before the evening easing begins.\par\smallskip
\textbf{Forecast instruction.} [720, 781) shows a sustained afternoon climb that ends close to the daily peak, after which demand levels off and turns down. Overall, under sunny skies, output reaches roughly its usual peak level.

\tcbsubtitle{F03 \enspace Traffic \enspace (News Forecast, California traffic)}\nopagebreak
\emph{Event-conditioned forecasting; Hourly resolution, history 120 and horizon 48 points; ROI [120, 143), 00:00 to 22:00 on 24 November 2016.}\par\nopagebreak\smallskip
\noindent\includegraphics[width=\linewidth]{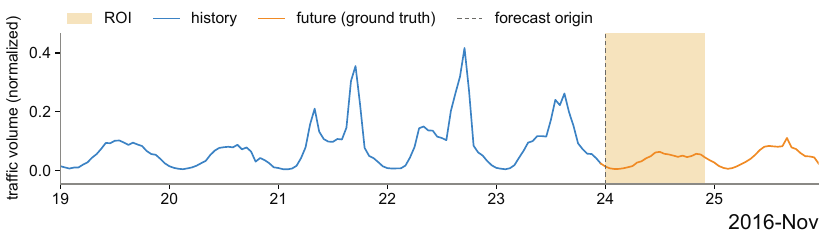}\par\smallskip
\textbf{Background.} Hourly traffic volume on a California freeway segment. The dominant patterns are a daily commute cycle (morning and evening peaks) and a weekly cycle (weekdays heavier than weekends); public holidays reshape the weekday pattern.\par\smallskip
\textbf{Event.} Thanksgiving Day, a US public holiday, falls on 24 November 2016 within the future window. The accompanying regional news headlines (16-17 November) contain no road-specific event.\par\smallskip
\textbf{Reasoning trace.} The relevant span is 24 November 00:00 through 22:00, Thanksgiving Day inside the future window. In the history window this freeway follows a stronger weekday commute cycle, but the holiday note points to a weekend-like profile instead: a muted build through the morning, a broad high around midday, and an easing later in the day rather than a pronounced commuter pattern.\par\smallskip
\textbf{Forecast instruction.} [120, 143) shows a gradual daytime rise to a midday crest followed by an uneven decline through the evening. Overall, on the holiday, the midday crest stays far below the usual weekday peaks.

\tcbsubtitle{F04 \enspace Finance \enspace (MTBench, US equities)}\nopagebreak
\emph{Event-conditioned forecasting; Intraday resolution, history 389 and horizon 78 points; ROI [389, 406), 17:19 to 18:36 UTC on 21 July 2022.}\par\nopagebreak\smallskip
\noindent\includegraphics[width=\linewidth]{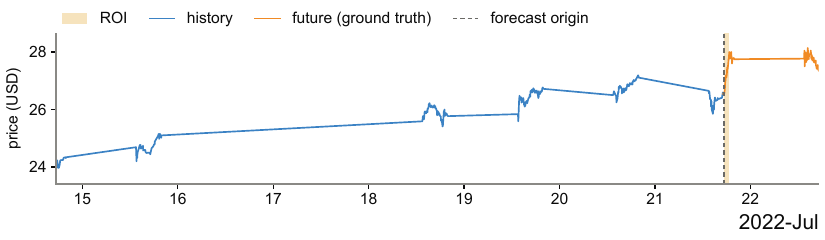}\par\smallskip
\textbf{Background.} Intraday price of a single US-listed stock over about one trading week of higher-frequency points, in US dollars. Moves are driven by company news, earnings and broad market sentiment, with no within-window seasonality.\par\smallskip
\textbf{Event.} On 21 July 2022 at 17:15 UTC, Dorchester Minerals, L.P. (Dallas, Texas) announced its second-quarter 2022 cash distribution, covering the three months ended 30 June 2022 and payable on 11 August 2022 to unitholders of record as of 1 August 2022.\par\smallskip
\textbf{Reasoning trace.} The selected span runs from 17:19 to 18:36 on 21 July, the first 75 minutes after the 17:15 announcement, so it holds the initial reaction to the cash distribution. A cash distribution is positive news for unitholders, and the last hours of the history window are flat, so the reaction is an advance from a calm base. That base sits near the top of the week's range in the history window, so an advance carries the series close to or above its previous highs.\par\smallskip
\textbf{Forecast instruction.} In [389, 406), the future window shows a sustained upward run that ends at its highest point and levels off afterward. Overall, the run ends slightly above the series' previous highs.

\tcbsubtitle{F05 \enspace Climate \enspace (Time-MMD, US drought)}\nopagebreak
\emph{Event-conditioned forecasting; Weekly resolution, history 36 and horizon 12 points; ROI [36, 41), 27 October to 24 November 2015.}\par\nopagebreak\smallskip
\noindent\includegraphics[width=\linewidth]{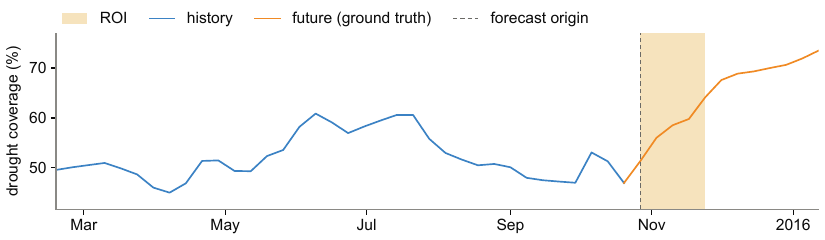}\par\smallskip
\textbf{Background.} Share of US land area under drought (US Drought Monitor categories D0-D4), in percent, reported weekly. It moves slowly and persistently: coverage builds over dry spells and recedes after sustained rain or snow, lagging precipitation by weeks.\par\smallskip
\textbf{Event.} Monthly precipitation reports for April-September 2015 (all before the horizon): April was the 37th wettest on record, May the wettest month ever recorded for the contiguous US (4.36 in), June the 9th wettest and July the 14th wettest (about 26\% of the US in moderate-to-exceptional drought as of 31 July). August turned dry (28th driest, 0.26 in below average) and the national drought footprint expanded; September was the 21st driest (0.40 in below average).\par\smallskip
\textbf{Reasoning trace.} The ROI spans 27 October (row 36) through 24 November 2015 (row 40), the first month of the future window. Both August and September were below average in precipitation and the drought footprint was already expanding at the close of August; coverage lags precipitation by weeks, so this deficit is due to appear in the weeks after the horizon. The history window does not show it yet: coverage fell from its July high to a local low at the horizon. Two consecutive dry months with the footprint already expanding point to a rise from that low sustained through the whole span, not a one-week uptick.\par\smallskip
\textbf{Forecast instruction.} The ROI [36, 41) captures a rapid expansion of drought coverage from late October through late November 2015, ending at its highest point in that span. Overall, after two dry months, coverage ends above the highest level of the history window.
\end{samplebox}

\subsection{Human review of the reasoning traces}
\label{app:audit}

\textbf{Sampling and review cards.} The 1{,}000 reviewed traces are a stratified sample drawn with a fixed seed. It holds 60 traces per domain in the understanding task (720) and 56 per domain in the forecasting task (280). Each trace is rendered on a review card with the question, the annotated answer and the trace. The card also shows the ground truth recomputed from the raw arrays: per-channel statistics and an evenly downsampled time series. For forecasting it further shows the history, the future and the values inside the ROI.

\textbf{Rubric.} Each trace receives one verdict, sound, minor or unsound. Unsound requires a definite defect: a statement that directly conflicts with the data, a conclusion that contradicts the annotated answer, or a fabricated fact. Three criteria are scored for the understanding task: U1, the trace does not contradict the time series; U2, the trace supports the annotated answer; U3, the trace contains no fabricated fact. Five criteria are scored for the forecasting task: F1, the ROI is localized inside the horizon; F2, the reading of the history is faithful; F3, the conclusion is consistent with the realized future; F4, the link from the event to its effect is plausible; F5, no value from the future window is cited.

\textbf{Reviewers and agreement.} The two reviewers are doctoral students with at least one year of research experience in time series processing, forecasting, or multimodal time series analysis. Each reviewer rates all 1{,}000 traces in a separate round with the same cards and rubric, and the second reviewer has no access to any first-round result. Table~\ref{tab:app-irr} reports the pass rate of each round, and the share of traces on which the two rounds give the same pass or fail label. The pass rates in Section~\ref{sec:dataset-review} are the means of the two rounds.

\begin{table}[h]
\centering
\footnotesize
\setlength{\tabcolsep}{6pt}
\caption{Human review of the reasoning traces. Agreement$_\mathrm{pass}$: share of traces on which the two review rounds give the same pass or fail label. Pass R1 / R2: pass rate of each round, in \%.}
\label{tab:app-irr}
\begin{tabular}{lccc}
\toprule
Scope & Agreement$_\mathrm{pass}$ & Pass R1 & Pass R2 \\
\midrule
All & 0.966 & 94.4 & 94.6 \\
Understanding & 0.957 & 93.1 & 93.5 \\
Forecasting & 0.989 & 97.9 & 97.5 \\
\bottomrule
\end{tabular}
\end{table}

\subsection{Detailed experimental setup}
\label{app:hyper}

\textbf{Stages and hardware.} Training has two stages. Stage one is supervised fine-tuning with the losses of Section~\ref{sec:setup}: text cross-entropy on the trace and the answer or instruction for both tasks, and the quantile (pinball) loss on the forecast for forecasting samples. Stage two applies GRPO \citep{grpo} on the understanding task. All stages run on NVIDIA H200 GPUs. Supervised fine-tuning takes ten epochs over the full understanding task on eight GPUs (49 h). GRPO runs on eight GPUs (25 h). Forecasting fine-tuning takes thirty epochs on two GPUs (5.5 h).

\textbf{Architecture settings and trainable parameters.} The planning rule of Appendix~\ref{app:swqformer} uses $W_{\min}=3$, $W_{\max}=23$, $k=8$ queries per window and $Q_{\mathrm{g}}=16$ global queries, with $W$ reaching $W_{\max}$ at $N=32{,}768$. Both Q-Formers stack two blocks of eight heads at width 768; the feedback Q-Former outputs 16 tokens. Training updates 240.7M parameters: 21.6M of LoRA on the LLM (rank 8, $\alpha$ 32, dropout 0.1, on every linear projection of the attention and MLP blocks, with embeddings and output head frozen), 147.9M on the TSFM side (the pretrained forecaster and the injected gated cross-attention), and 48.5M and 22.7M in the sliding-window and the feedback Q-Former. The gates start at 1.0 ($\tanh \approx 0.76$) and the cross-attention output projections start at zero, so at initialization the TSFM computes the same function as the pretrained model.

\textbf{Optimization and batching.} Four parameter groups are used: $10^{-5}$ for the TSFM, $10^{-4}$ for the LoRA adapter, $10^{-4}$ for the Q-Formers and the injected cross-attention projections, and $10^{-3}$ for the gates, which carry no weight decay. Understanding and forecasting share this setting. The optimizer is AdamW with linear decay and 3\% warmup, no weight decay, and gradient clipping at 1.0, under bf16, gradient checkpointing, and ZeRO-2 data parallelism. Batches are packed dynamically per device up to a padded-token budget, where the cost of a sample is its text tokens plus its planned soft-prompt tokens. The budget is 155k tokens for forecasting and 20k for understanding.

\label{app:ssrl}
\textbf{Scheduled sampling.} On forecasting batches, the teacher-forcing probability of the feedback text \citep{scheduledsampling} decays linearly with the training step from 1.0 to 0.7, so that late in training 30\% of batches feed the model's own generated text back to the forecaster. The draw is made once per batch and synchronized across devices. Self-generated text is decoded greedily with at most 288 new tokens and carries no text loss: the cross-entropy is computed on the teacher-forced pass in every batch.

\textbf{GRPO.} Stage two samples $G=8$ traces per question at temperature 1.0. The reward is exact match of the final answer, and the intermediate reasoning earns no reward. No KL penalty is applied. Advantages are standardized within each group. GRPO updates only the LoRA adapter of the LLM (21.6M parameters) with learning rate $10^{-5}$; all other parameters are frozen. Questions are sampled by domain, with a higher rate for the domains with fewer training samples. Each step draws 64 questions across devices, giving 512 rollouts of at most 2{,}048 tokens; the update is a single on-policy pass per group, with no reference model and no importance clipping.

\textbf{Generation budget of the baselines.} For each baseline that decodes its output as text, the generation budget starts from the default of its official code and is adjusted to the output lengths that \dataset{} requires, so that a complete answer or forecast fits within it. A baseline whose code sets no default is given the same budget as \model{}.

\label{app:sr}

\textbf{Scoring of the forecasts.} A model that outputs one point trajectory has that trajectory copied to all 21 quantile levels, which reduces its CRPS to a weighted absolute error. Models with a single-channel interface read the first channel of a multivariate sample, whereas \model{} reads all channels. Forecasting scores are computed per sample and then averaged. A model that decodes its forecast as text can fail to produce a parseable array, and an average over the parsed samples alone leaves these failures out of the score. The parsed samples are also the easier ones: for the four rows with parse failures, the median horizon of the failed samples is 63 to 96 steps, against 24 to 48 for the parsed samples (Table~\ref{tab:app-sr}). The scores of Table~\ref{tab:main-fc} are therefore weighted per domain $d$ by the success rate
\begin{gather}
\mathrm{sr}_d=\frac{\#\{\text{samples of } d \text{ with a finite forecast over the full horizon}\}}{\#\{\text{test samples of } d\}},\\
\mathrm{CRPS}_d=\frac{\mathrm{CRPS}_d^{\text{parsed}}}{\mathrm{sr}_d},
\end{gather}
and Avg.\ is the equal-weight mean of the five weighted domain scores. The full-horizon and the ROI scores share $\mathrm{sr}_d$, since parsing is a property of the generation and not of the scored region. The weight is applied per domain because failures concentrate in the domains with the largest CRPS: the zero-shot LLM parses 46.2\% of all samples but 27.6\% of Solar (Table~\ref{tab:app-fc-sr-domain}). Only the four rows of Table~\ref{tab:app-sr} have $\mathrm{sr}_d<1$. The other rows of Table~\ref{tab:main-fc}, including \model{}, whose forecast is the TSFM's numeric output and involves no parsing, have $\mathrm{sr}_d=1$ on all five domains and are unchanged by the weighting.

\begin{table}[h]
\centering
\small
\caption{The four rows of Table~\ref{tab:main-fc} with parse failures. \textbf{Success rate}: share of the test samples with a parsed forecast; Table~\ref{tab:app-fc-sr-domain} gives it per domain. \textbf{Horizon}: median horizon of the parsed / failed samples. \textbf{CRPS}: Avg.\ over the parsed samples alone, and after weighting (the value of Table~\ref{tab:main-fc}).}
\label{tab:app-sr}
\begin{tabular}{lcccc}
\toprule
 & & Horizon & \multicolumn{2}{c}{CRPS} \\
\cmidrule(lr){3-3}\cmidrule(lr){4-5}
Model & Success rate (\%) & parsed / failed & parsed & weighted \\
\midrule
\model{}'s LLM & 46.2 & 24 / 96 & 0.302 & 0.726 \\
\model{}'s LLM$^\dagger$ & 82.8 & 48 / 96 & 0.238 & 0.308 \\
Time-MQA-7B & 71.4 & 48 / 96 & 0.671 & 0.863 \\
TimeOmni-1-4B & 80.8 & 48 / 63 & 0.365 & 0.415 \\
\bottomrule
\end{tabular}
\end{table}

\subsection{Additional results}
\label{app:per-dataset}

\textbf{SciTS by domain.} Table~\ref{tab:app-scits} expands Figure~\ref{fig:ood}a to all ten domains.

\begin{table}[h]
\centering
\footnotesize
\setlength{\tabcolsep}{1.5pt}
\caption{SciTS results by domain. Per-domain task counts are 2/3/1/1/3/2/4/4/2/1 (23 tasks over 22 datasets). \model{} is trained on the official training split; the other rows are transcribed from SciTS. MIMII is excluded because its training split contains no positive sample, so every row's Manuf.\ score is recomputed without it from the per-task results of SciTS. $(x/y)$: ChatTS-14B completes only $x$ of the domain's $y$ tasks, so its average is undefined. Bold marks the best score per column.}
\label{tab:app-scits}
\begin{tabular}{lccccccccccc}
\toprule
Model & Astron. & Bioac. & Earth. & Econ. & Meteo. & Manuf. & Neuro. & Phys. & Radar & Urban. & \textbf{Avg.} \\
\midrule
\multicolumn{12}{l}{\emph{General-purpose LLMs}} \\
Gemini2.5-Flash (text) & 40.2 & 10.3 & 67.6 & 87.8 & 51.8 & 9.8 & 12.7 & 31.8 & 17.2 & 64.6 & 39.4 \\
GPT-5-mini & 42.3 & 10.7 & 67.6 & 83.8 & 45.3 & 39.0 & 13.9 & 25.0 & 16.5 & 64.8 & 40.9 \\
Gemini2.5-Flash (multimodal) & 38.4 & 7.7 & 72.5 & 82.9 & 52.3 & 32.1 & 29.8 & 27.3 & 21.4 & 67.0 & 43.1 \\
\multicolumn{12}{l}{\emph{Time series models}} \\
ChatTS-14B & 11.3 & (0/3) & 64.8 & 79.2 & 51.2 & 0.0 & 22.7 & 30.9 & 13.9 & 65.4 & -- \\
UniTS & 38.2 & 8.1 & 0.0 & 27.1 & 9.8 & 39.4 & 25.9 & 22.9 & 10.6 & \textbf{67.4} & 24.9 \\
TimeOmni (the suite's own) & 73.2 & 58.1 & 82.5 & 96.4 & 61.3 & 92.5 & 60.1 & 45.9 & 68.9 & 64.8 & 70.4 \\
\midrule
\model{} & \textbf{94.3} & \textbf{67.0} & \textbf{100.0} & \textbf{99.0} & \textbf{67.1} & \textbf{100.0} & \textbf{68.7} & \textbf{73.9} & \textbf{76.2} & \textbf{67.4} & \textbf{81.4} \\
\bottomrule
\end{tabular}
\end{table}

\textbf{Forecasting scores and parse success of all models.} Table~\ref{tab:app-fc-sr-domain} extends Table~\ref{tab:main-fc} to the ROI. It also separates the two quantities that the success-rate weighting (Appendix~\ref{app:sr}) combines: whether a forecast is parsed, and the error of the parsed forecasts. CRPS is computed over the parsed samples alone, so it equals the value of Table~\ref{tab:main-fc} for the nine rows without parse failures.

\begin{table}[h]
\centering
\footnotesize
\setlength{\tabcolsep}{1.8pt}
\caption{Forecasting by domain for all rows of Table~\ref{tab:main-fc}: CRPS over the parsed samples, as full horizon / ROI, lower is better. The four rows with parse failures carry their success rate on a second line, and all other rows parse every forecast. \textbf{Avg.} is the mean of the five domains for CRPS and the share of all test samples for the success rate. Dividing a CRPS by its success rate gives the weighted score of Table~\ref{tab:main-fc}. $\dagger$: fine-tuned on the \dataset{} training split.}
\label{tab:app-fc-sr-domain}
\begin{tabular}{@{}lcccccc@{}}
\toprule
Model & Solar & Load & Traffic & Finance & Climate & \textbf{Avg.} \\
\midrule
ChatTime-1-7B & 0.898 / 0.804 & 0.106 / 0.113 & 0.376 / 0.458 & 0.058 / 0.053 & 0.107 / 0.089 & 0.309 / 0.303 \\
Time-MQA-7B & 2.252 / 1.258 & 0.197 / 0.207 & 0.719 / 0.733 & 0.098 / 0.087 & 0.089 / 0.081 & 0.671 / 0.473 \\
\quad\emph{success rate (\%)} & \emph{78.5} & \emph{54.5} & \emph{89.0} & \emph{60.3} & \emph{76.9} & \emph{71.4} \\
TimeReasoner & 0.640 / 0.602 & 0.081 / 0.084 & 0.271 / 0.268 & 0.041 / 0.040 & 0.110 / 0.100 & 0.229 / 0.219 \\
DoubleCast & 0.530 / 0.470 & 0.083 / 0.109 & 0.180 / 0.201 & 0.029 / 0.026 & 0.065 / 0.054 & 0.177 / 0.172 \\
TimeOmni-VL-15B & 4.597 / 2.108 & 0.466 / 0.408 & 1.305 / 1.252 & 0.101 / 0.098 & 0.218 / 0.204 & 1.337 / 0.814 \\
TimeOmni-1-4B & 0.951 / 0.840 & 0.132 / 0.149 & 0.562 / 0.596 & 0.049 / 0.043 & 0.132 / 0.115 & 0.365 / 0.349 \\
\quad\emph{success rate (\%)} & \emph{95.1} & \emph{51.2} & \emph{89.0} & \emph{86.2} & \emph{100} & \emph{80.8} \\
UniTS-x128$^\dagger$ & 0.865 / 0.670 & 0.059 / 0.063 & 0.276 / 0.311 & 0.033 / 0.028 & 0.111 / 0.093 & 0.269 / 0.233 \\
TimeOmni$^\dagger$ & 0.468 / 0.414 & 0.042 / 0.047 & 0.224 / 0.288 & 0.033 / 0.029 & 0.085 / 0.066 & 0.170 / 0.169 \\
\midrule
\model{}'s LLM & 0.629 / 0.615 & 0.181 / 0.194 & 0.410 / 0.490 & 0.072 / 0.063 & 0.217 / 0.191 & 0.302 / 0.311 \\
\quad\emph{success rate (\%)} & \emph{27.6} & \emph{56.2} & \emph{65.9} & \emph{37.9} & \emph{100} & \emph{46.2} \\
\model{}'s LLM$^\dagger$ & 0.779 / 0.776 & 0.061 / 0.073 & 0.212 / 0.205 & 0.040 / 0.036 & 0.097 / 0.077 & 0.238 / 0.233 \\
\quad\emph{success rate (\%)} & \emph{69.9} & \emph{86.0} & \emph{100} & \emph{84.5} & \emph{100} & \emph{82.8} \\
\model{}'s TSFM & 0.511 / 0.451 & 0.066 / 0.072 & 0.187 / 0.231 & 0.032 / 0.029 & 0.065 / 0.058 & 0.172 / 0.168 \\
\model{}'s TSFM$^\dagger$ & 0.378 / 0.341 & 0.037 / 0.037 & 0.138 / 0.147 & 0.025 / 0.022 & 0.065 / 0.063 & 0.129 / 0.122 \\
\midrule
\textbf{\model{} (Ours)} & \textbf{0.357} / \textbf{0.317} & \textbf{0.032} / \textbf{0.034} & \textbf{0.113} / \textbf{0.110} & \textbf{0.022} / \textbf{0.021} & \textbf{0.056} / \textbf{0.052} & \textbf{0.116} / \textbf{0.107} \\
\bottomrule
\end{tabular}
\end{table}

\label{app:source-papers}
\textbf{Against the source papers.} Table~\ref{tab:source-papers} compares the single \model{} checkpoint against the best results \emph{reported by the source datasets' own papers}, on the subtasks where the protocols match: same test items, same input candidate space and same scoring. Every compared model is specialized and trained on that single dataset (OpenTSLM's curriculum-trained 1--3B models, STReasoner-8B with spatial-aware RL, VeriTime's 3--4B SFT+RL models), whereas \model{} is one 9.2B checkpoint trained on the understanding task of \dataset{} that additionally outputs a reasoning trace. \model{} outperforms the reported best on 9 of the 11 subtasks. The two losses are on the TSRBench scenario-attribution and anomaly-detection subtasks against VeriTime's SFT+RL-specialized model; the anomaly-detection loss is $-2.78$ and the scenario-attribution loss is $-7.39$.

\begin{table}[h]
\centering
\footnotesize
\setlength{\tabcolsep}{3pt}
\caption{\model{} (one checkpoint) vs.\ the best result reported by each source paper, on protocol-clean subtasks. Metric is accuracy unless noted.}
\label{tab:source-papers}
\begin{tabular}{llccc}
\toprule
Source paper & Subtask & \model{} & Best reported (model) & $\Delta$ \\
\midrule
OpenTSLM & ECG-QA & \textbf{57.95} & 46.25 (Flamingo-Llama3.2-3B) & $+11.70$ \\
ST-Bench & T1 cause & \textbf{97.58} & 95.65 (STReasoner-8B) & $+1.93$ \\
ST-Bench & T2 entity & \textbf{97.15} & 75.71 (STReasoner-8B) & $+21.44$ \\
ST-Bench & T3 correlation/causality & \textbf{95.85} & 87.12 (STReasoner-8B) & $+8.73$ \\
TSRBench & Anomaly detection & 88.33 & \textbf{91.11} (VeriTime-4B) & $-2.78$ \\
TSRBench & Scenario attribution & 80.11 & \textbf{87.50} (VeriTime-4B) & $-7.39$ \\
TSRBench & Inferential calculation & \textbf{81.90} & 77.14 (VeriTime-4B) & $+4.76$ \\
TSRBench & CTU & \textbf{77.08} & 67.50 (VeriTime-4B) & $+9.58$ \\
TSRBench & ECG & \textbf{36.50} & 30.30 (VeriTime-4B) & $+6.20$ \\
TSRBench & EMG & \textbf{87.33} & 64.96 (VeriTime-3B) & $+22.37$ \\
TSRBench & RCW & \textbf{78.50} & 64.89 (VeriTime-3B) & $+13.61$ \\
\bottomrule
\end{tabular}
\end{table}

\subsection{Trace-quality evaluation: protocol}
\label{app:trace-eval}

\textbf{Samples and judge inputs.} On understanding, the judged set holds 844 test samples. Each domain contributes $2\sqrt{n}$ samples, clamped to $[40, 250]$ and stratified within the domain by subtask and by whether the model's answer is correct (525 correct, 319 wrong). The judge receives the question, the model's answer, the model's trace and the annotated trace, and no representation of the time series; both metrics are scored in one call (Prompt~\ref{box:judge-und}). On forecasting, all test samples are judged. Consistency is the mean of two judges. An instruction judge compares the generated forecast instruction with the annotated one on the shape of the horizon alone, ignoring index ranges, clock times and magnitude qualifiers (Prompt~\ref{box:judge-fc-instr}). A reasoning judge compares the generated reasoning with the annotated trace on the region of interest, the effect attributed to the event and the behaviour the history supports (Prompt~\ref{box:judge-fc-reas}). For supportiveness, the judge receives the full generated text together with the model's own forecast, rendered as a plot of the last 200 history steps, the median forecast with its 10--90\% quantile band and the ROI shaded, and listed as the median values in text; no ground truth is shown (Prompt~\ref{box:judge-fc-sup}). Every judge is GPT-5.4 through the chat API with default decoding, called once per sample and condition, and replies with a JSON object holding the score and a one-sentence reason.

\textbf{Mismatched reference.} The chance reference re-scores every trace against a partner sample. Partners come from a single-cycle derangement of the judged set, a random permutation without fixed points in which every sample serves as a partner exactly once, drawn once with a fixed seed. Exactly one field is exchanged and the prompts are unchanged. On understanding, consistency pairs the trace with the partner's annotated trace, and supportiveness scores the trace inside the partner's context, its question, answers and annotated trace. On forecasting, both consistency judges pair the generated text with the partner's annotated instruction and trace, using the same partner, and supportiveness pairs the partner's generated text with the sample's own forecast plot and values. The matched and mismatched scores of Table~\ref{tab:trace-eval} are means over the judged set; on forecasting, the consistency entry is the mean of the instruction judge and the reasoning judge under the same pairing.

\textbf{Supportiveness by answer correctness.} The matched supportiveness scores on understanding are split by whether the model's answer is correct, using the correctness labels of the judged set, and the two means are compared over the whole set and within each domain.

\textbf{Counterfactual magnitude edit.} The edit is applied to the annotated text of each forecasting test sample rather than to generated reasoning, so that the five variants of a sample differ in the magnitude statement alone. The sentence of the annotated forecast instruction that states the relative magnitude is replaced in turn by each of the five statements of Table~\ref{tab:app-cf-levels}, ordered from the lowest to the highest stated magnitude, while the rest of the instruction, the annotated trace, the event description and the history stay fixed. Each variant is teacher-forced through the LLM-to-TSFM pathway, so no text is generated, and the TSFM returns its quantile forecast. The peak of a variant is the maximum of the median forecast over the ROI. For each sample, Spearman's $\rho$ between the order of the five statements and the five peaks is computed and then averaged over samples; direction agreement is the share of samples whose peak under \emph{above} strictly exceeds the peak under \emph{near zero}. The chance reference shuffles each sample's five peaks across the statements 1{,}000 times and computes the same two quantities on the shuffled peaks.

\begin{table}[h]
\centering
\footnotesize
\caption{The magnitude statements substituted into the forecast instruction in the counterfactual edit, ordered from the lowest to the highest stated magnitude.}
\label{tab:app-cf-levels}
\begin{tabular}{ll}
\toprule
Level & Magnitude statement \\
\midrule
Near zero & Overall, the peak reaches only a small fraction of the series' typical high level. \\
Well below & Overall, the peak runs well below the series' typical highs. \\
Moderately below & Overall, the peak sits moderately below the series' typical highs. \\
Close to & Overall, the peak sits close to the series' typical high level. \\
Above & Overall, the peak runs well above the series' typical highs. \\
\bottomrule
\end{tabular}
\end{table}

\begin{promptbox}[label={box:judge-und}]{Understanding judge, consistency and supportiveness in one call}
System message:

You are a strict evaluator of explanations produced by a time-series analysis model. The model sees the time series ONLY through a learned encoder — it has NO access to the raw numeric values as text. Reply with a single JSON object and nothing else.

User message:

Question posed to the model:
{question}

Model's answer: {pred_answer}

Model's explanation:
{explanation}

Reference reasoning (may use different wording):
{gt_reasoning}

Rate the model's EXPLANATION on two axes and reply with JSON:
{"support": <1-5>, "consistency": <1-5>, "reason": "<one sentence>"}

- "support": does the explanation actually justify THE ANSWER THE MODEL GAVE? Judge internal
  coherence only — do NOT reward or punish based on whether that answer is correct.
  1 = unrelated or self-contradictory, 5 = directly and specifically justifies it.
- "consistency": do the explanation's criteria match the reference reasoning's criteria?
  Ignore wording, ordering, and any numeric values. 1 = unrelated criteria, 5 = same criteria.
\end{promptbox}

\begin{promptbox}[label={box:judge-fc-instr}]{Forecasting instruction judge, first half of consistency}
System message:

You are a careful evaluator comparing two textual descriptions of a time series' future trajectory. Judge ONLY the qualitative SHAPE/trend narrative -- direction of change, number and location of turning points, smoothness vs. jaggedness, spikes, plateaus, oscillation. Explicitly IGNORE numeric index ranges, clock times/dates, and magnitude-level qualifiers (e.g. 'well below the series' typical highs') -- those are scored by a separate metric, not by you.

User message:

Compare the SHAPE of the future trajectory described in the two passages below.

[Model-generated text]
{gen_text}

[Reference (ground-truth) shape description]
{gt_conclusion}

Rate how semantically consistent the SHAPE described in the model-generated text is with the shape in the reference, on this 1-5 scale:
5 = same shape narrative (same direction(s) of change and turning-point structure)
4 = mostly consistent, minor difference in nuance/degree but same overall narrative
3 = partially consistent -- same broad direction but a clearly different shape detail (e.g. smooth vs. jagged, or a different number of turning points)
2 = mostly inconsistent -- different overall shape narrative, only superficial overlap
1 = contradictory / opposite shape (e.g. rises vs. falls, flat vs. spiking)

Respond with ONLY a JSON object of the form {"score": <int 1-5>, "reason": "<one short sentence>"}
\end{promptbox}

\begin{promptbox}[label={box:judge-fc-reas}]{Forecasting reasoning judge, second half of consistency}
System message:

You are a careful evaluator comparing two analytical reasoning passages written about the same time-series forecasting task. Judge ONLY whether the model's reasoning reaches the same substantive analysis as the reference: which part of the future window is identified as the region of interest, how the stated event is expected to affect it, and what behaviour the history is said to support. Explicitly IGNORE writing style, ordering, verbosity, and exact clock times / index numbers.

User message:

Compare the substantive analysis in the two passages below.

[Model-generated reasoning]
{gen_text}

[Reference (ground-truth) reasoning]
{gt_text}

Rate how substantively consistent the model's reasoning is with the reference, on this 1-5 scale:
5 = same analysis -- same region of interest, same attributed effect of the event, same expected behaviour
4 = mostly consistent -- same core analysis, minor difference in emphasis or one secondary detail
3 = partially consistent -- overlaps on the broad situation but differs on a substantive point (a different expected effect, or a clearly different region)
2 = mostly inconsistent -- different analysis, only superficial overlap (shared domain vocabulary)
1 = contradictory -- asserts an effect or a region opposite to the reference

Respond with ONLY a JSON object of the form {"score": <int 1-5>, "reason": "<one short sentence>"}
\end{promptbox}

\begin{promptbox}[label={box:judge-fc-sup}]{Forecasting supportiveness judge}
System message:

You are a careful evaluator checking whether a numeric time-series forecast (shown as a plot AND as exact numbers) is CONSISTENT with a separate piece of reasoning text that was supposed to describe/justify that same forecast. You are NOT checking whether the forecast is accurate against any ground truth -- there is no ground truth here. You are ONLY checking whether the forecast (what the model actually predicted) matches what the reasoning text says it predicted: direction of change, magnitude/level qualifiers, shape (peaks, troughs, turning points, smoothness), and any claims specific to the highlighted ROI region.

User message:

[Reasoning text generated by the model]
{gen_text}

[The forecast the model actually produced]
The plot below shows recent history (gray) followed by the model's own predicted forecast (blue dashed = median, shaded band = 10-90% interval). The shaded yellow region is the ROI (region of interest) that the reasoning text specifically discusses.

Exact predicted median values for the forecast window (rounded, chronological order): {values_str}
ROI covers steps [{roi_start}, {roi_end}) within this window; predicted median values there: {roi_values_str}

Rate how CONSISTENT the actual forecast (plot + numbers above) is with what the reasoning text describes, on this 1-5 scale:
5 = fully consistent -- direction, magnitude, and shape all match the text
4 = mostly consistent -- minor nuance mismatch but same overall story
3 = partially consistent -- right broad direction but a clearly wrong detail (e.g. wrong shape, or magnitude claim not supported by the numbers)
2 = mostly inconsistent -- forecast contradicts a central claim of the text
1 = contradictory -- forecast is essentially the opposite of what the text describes

Respond with ONLY a JSON object of the form {"score": <int 1-5>, "reason": "<one short sentence>"}
\end{promptbox}

\subsection{Case studies}
\label{app:case}

This section presents two understanding cases and two forecasting cases; the statistics over the full test sets are those of Section~\ref{sec:results-ablation}. Baselines are scored leniently throughout: an output counts as correct if the ground-truth phrase appears anywhere in it, a bare option letter counts, and the last number of a numeric answer is taken. In every quoted trace, the cited values are approximate, because the LLM reads the time series through the soft prompt rather than as digits; what is checked against the raw time series is the structural judgment: how many events, how large, and which shape.

\begin{figure}[!htb]
\centering
\includegraphics[width=\linewidth]{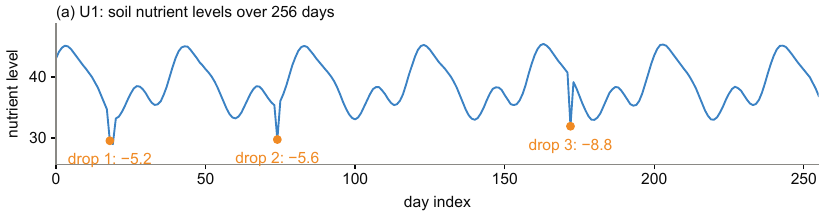}\\[2pt]
\includegraphics[width=\linewidth]{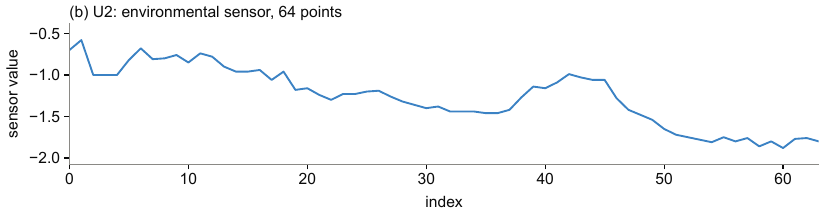}
\caption{The two understanding cases. \textbf{(a)} U1: the three largest one-day drops, which the trace counts. \textbf{(b)} U2: a gradual decline with one partial rebound and no jump, spike or change of variance.}
\label{fig:case-u}
\end{figure}

\textbf{Case U1, counting drops (TSRBench, Earth science).} The question asks how many significant downward spikes in a 256-day soil-nutrient time series could critically impact plant growth; the answer is 3 (Figure~\ref{fig:case-u}a). The trace of \model{} sets a criterion first and then enumerates: it lists three drops of about 7 to 8 units with their days, and it names smaller declines that it rules out. Against the raw time series, exactly three one-day drops exceed 5 units (5.2, 5.6 and 8.8) and the fourth largest is 1.3. Every other model that produces a number is off by one (Table~\ref{tab:app-case-und}): the no-reasoning model and TimeOmni-1 answer 2, the fine-tuned LLM and TimeOmni answer 1, and Time-MQA detects only the first drop. The models that state a count directly undercount, whereas the trace counts the drops one by one against a stated criterion.

\textbf{Case U2, fine-grained diagnosis (RATs40K, Earth science).} The 21-way question asks which category describes a 64-point environmental-sensor time series; the answer is normal sequence (Figure~\ref{fig:case-u}b). The time series declines from $-0.70$ to $-1.80$ with one partial rebound around index 40 and no jump, spike or change of variance; in the source taxonomy a drift anomaly requires a systematic departure from a baseline, not a gradual decline. Every other model that produces a category reports drift (the no-reasoning model, the fine-tuned LLM, TimeOmni, TimeOmni-1 and Time-MQA), and TimeOmni-VL reports trend divergence. The trace of \model{} checks for disruptions, spikes and irregularities first and assigns the category only then. The same pattern holds over the 453 univariate RATs40K test samples: 188 are normal, the no-reasoning model labels 132 of them normal and predicts drift 96 times against 58 true, and \model{} labels 177 normal and predicts drift 62 times.

\begin{table}[!htb]
\centering
\footnotesize
\setlength{\tabcolsep}{4pt}
\caption{Answers of every compared model on the two understanding cases, scored leniently. Bold marks a correct answer. ``No answer'': the generation budget ended inside the reasoning. ``Method only'': the output lists steps without a result. ``Degenerate'': token dumps, a bare option letter, or a numeric continuation.}
\label{tab:app-case-und}
\begin{tabular}{lll}
\toprule
Model & U1 (count) & U2 (category) \\
\midrule
Ground truth & 3 & normal sequence \\
\midrule
\textbf{\model{} (Ours)} & \textbf{3} & \textbf{normal sequence} \\
No-reasoning & 2 & drift anomaly \\
\model{}'s LLM & no answer & no answer \\
\model{}'s LLM$^\dagger$ & 1 & drift anomaly \\
TimeOmni$^\dagger$ & 1 & drift anomaly \\
TimeOmni-1-4B & 2 & drift anomaly \\
Time-MQA-7B & 1 & drift anomaly \\
TimeOmni-VL-15B & method only & trend divergence anomaly \\
Chat-TS-8B, ChatTime-1-7B & degenerate & degenerate \\
\bottomrule
\end{tabular}
\end{table}

\begin{figure}[!htb]
\centering
\includegraphics[width=\linewidth]{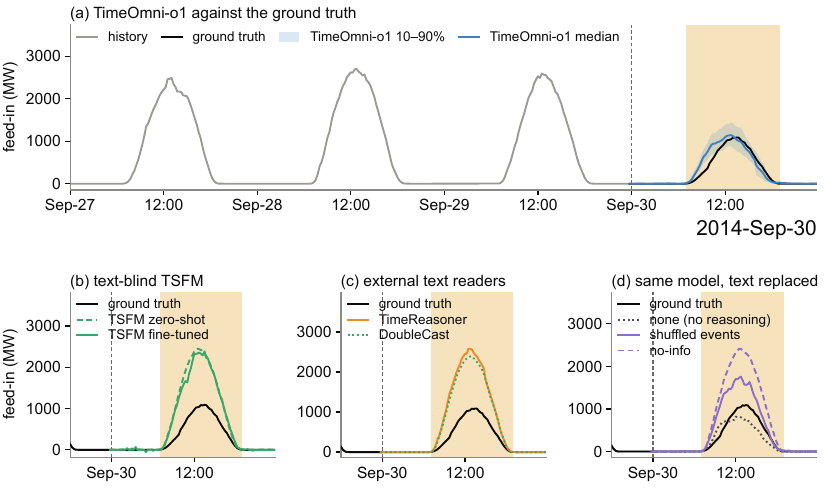}
\caption{Case F1, solar feed-in in the 50Hertz zone after three days of high output, with rain in the forecast. \textbf{(a)} \model{} follows the stated magnitude and matches the low day. \textbf{(b)} The text-blind TSFM forecasts a fourth day at the previous level. \textbf{(c)} The time series models that read the text forecast the previous level as well. \textbf{(d)} The same checkpoint reverts to the previous level when the text is uninformative; the no-reasoning model lowers the peak, but its forecast falls after the peak as slowly as it rose. The shaded span is the annotated ROI.}
\label{fig:case-f1}
\end{figure}

\begin{figure}[!htb]
\centering
\includegraphics[width=\linewidth]{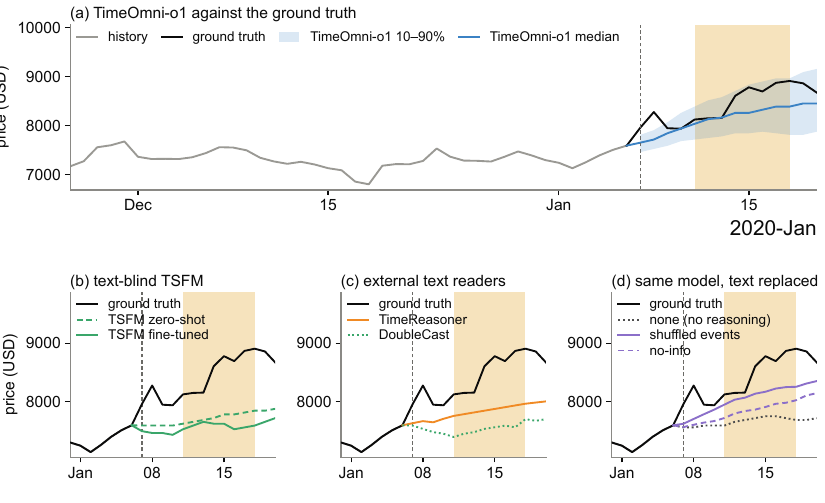}
\caption{Case F2, Bitcoin over 14 days after bullish halving headlines. \textbf{(a)} \model{} rises from the first day. \textbf{(b)} The fine-tuned TSFM stays near the origin; the zero-shot one rises too little. \textbf{(c)} TimeReasoner reads the momentum and forecasts conservatively. \textbf{(d)} The no-reasoning model stays flat on the same headlines.}
\label{fig:case-f2}
\end{figure}

The two forecasting cases share one figure layout (Figures~\ref{fig:case-f1} and~\ref{fig:case-f2}). Panel (a) shows the full history, the ground truth and the median forecast of \model{} with its 10 to 90\% band. Panels (b) to (d) show the horizon only and compare the ground truth with the text-blind TSFM (zero-shot and fine-tuned), with the two time series models that read the text (TimeReasoner and DoubleCast), and with the same checkpoint under the three text conditions of Figure~\ref{fig:abl}(b): no-reasoning, shuffled events and no-info.

\textbf{Case F1, the weather determines the magnitude (Fidel-TS 50Hertz, Solar).} After three days with daily peaks of 2{,}490 to 2{,}700 MW, the text for the forecast day reads: a mix of sunshine and clouds, light rain showers in the late afternoon, a chance of fog in the evening. The realized peak is 1{,}090 MW. The trace assigns the event to the daytime window, derives a low, broad peak from the mixed sky and a faster decline after it from the showers, and the instruction states a magnitude well below the typical peak. The forecast conforms to the instruction (Figure~\ref{fig:case-f1}a): its peak is 1{,}144 MW. Every model that does not read the text, or reads it without acting on it, forecasts a fourth day at the previous level. The zero-shot and fine-tuned TSFM peak at 2{,}448 and 2{,}384 MW and DoubleCast at 2{,}390; TimeReasoner peaks at 2{,}581 although its reasoning quotes the sun, clouds and rain and then judges that ``the average is fine''. Panel (d) compares the text conditions on one checkpoint. With an uninformative text the forecast reverts to the profile of the previous days with a peak of 2{,}414 MW, the level of the zero-shot TSFM; with another day's weather the peak is 1{,}758 MW. The no-reasoning model reads the same text and lowers the peak to 820 MW, below the realized peak. Its forecast falls after the peak as slowly as it rose: 0.86 of the peak two hours before and after. The truth and \model{} fall faster after the peak (0.79 to 0.66 and 0.85 to 0.70), the decline that the trace derives from the showers.

\textbf{Case F2, the news determines the direction (News Forecast, Bitcoin).} The history closes at 7{,}588 after a sustained rise. The headlines before the origin carry halving momentum toward 7{,}600 and a supportive macro report, and two negatives, a Bitmain layoff report and regulatory pressure from Turkey. The truth climbs to 8{,}907 and closes at 8{,}670. The trace weighs the four items, treats the two negatives as a short-term offset and the halving momentum as the dominant factor; the instruction states a sharp jump to a new high followed by a steady climb. The forecast rises from the first day and ends at 8{,}448. The fine-tuned TSFM dips to 7{,}424 and ends at 7{,}712, near the origin. The zero-shot TSFM and TimeReasoner rise in the right direction but only to 7{,}872 and 8{,}000; the reasoning of TimeReasoner reads the momentum and then forecasts ``about \$7900 to \$8000''. The no-reasoning model reads the same headlines and stays almost flat at 7{,}744, whereas the trace resolves the conflicting headlines into a direction.

\subsection{Architecture ablations}
\label{app:arch-abl}

\textbf{Scope.} The ablations of Section~\ref{sec:results-ablation} vary what the model reads; those reported here vary how it is built. To keep every arm and its baseline on one recipe, all ablation arms in this section are trained on the carrier subtask alone, so their scores are not those of the \dataset{} checkpoints of Section~\ref{sec:results-main} and are not comparable with them. Six choices are tested. On the series-to-LLM pathway: the form of the sliding-window Q-Former, the global queries, and the TSFM block that feeds the soft prompt. On the LLM-to-TSFM pathway: the form of the feedback Q-Former, the LLM layer it reads, and the set of TSFM blocks it injects into.

\textbf{Carriers.} The Q-Former ablations use ECG-QA, the understanding subtask with the longest time series: its 12-lead, 5{,}000-step recordings yield 3{,}756 patches per sample, reduced 22$\times$ to 168 soft-prompt tokens.

\textbf{Protocol.} An architectural change cannot resume from the \dataset{} checkpoint, so each understanding arm is trained from the pretrained components on its carrier alone. ECG-QA arms train on a 60k-sample subset of its training split for four epochs and are scored on a fixed 4{,}000-sample test subset. Forecasting arms replicate the recipe of the Table~\ref{tab:main-fc} checkpoint exactly, changing only the ablated switch, and are scored under its protocol (equal-weight domain mean, model-generated reasoning).

\begin{table}[h]
\centering
\footnotesize
\caption{Architecture ablations. \textbf{Top:} series-to-LLM pathway on ECG-QA, accuracy on a fixed 4{,}000-sample subset. \textbf{Bottom:} LLM-to-TSFM pathway, CRPS over the full horizon under the protocol of Table~\ref{tab:main-fc}. $\Delta$ is relative to \model{}.}
\label{tab:app-arch}
\setlength{\tabcolsep}{5pt}
\begin{tabular}{lcc}
\toprule
Series to LLM (ECG-QA) & Acc. & $\Delta$ \\
\midrule
\model{} & 53.33 & -- \\
Sliding-window Q-Former $\to$ mean pooling & 52.45 & $-0.88$ \\
Global queries $Q_{\mathrm{g}}{=}16 \to 0$ & 52.68 & $-0.65$ \\
Soft prompt from block 12 $\to$ 6 & 52.85 & $-0.48$ \\
\bottomrule
\end{tabular}

\vspace{4pt}
\begin{tabular}{lcc}
\toprule
LLM to TSFM (forecasting) & CRPS & $\Delta$ \\
\midrule
\model{} & 0.116 & -- \\
Feedback Q-Former $\to$ mean pooling & 0.126 & $+0.010$ \\
LLM layer 32 $\to$ 24 & 0.133 & $+0.017$ \\
Inject into 12 blocks $\to$ last 6 & 0.133 & $+0.017$ \\
Inject into 12 blocks $\to$ first 6 & 0.134 & $+0.018$ \\
\bottomrule
\end{tabular}
\end{table}

\textbf{Results.} Table~\ref{tab:app-arch} lists the results. On the series-to-LLM pathway, all three variants stay within 0.9 points of \model{}. Dropping the global queries raises $W_{\min}$ and $W_{\max}$ to 5 and 25, so on ECG-QA the window count rises from 19 to 21 and the soft prompt keeps 168 tokens; only the global branch is removed. On the LLM-to-TSFM pathway, the four variants trail \model{} by 0.010 to 0.018 CRPS, the same order as the gain of \model{} over the fine-tuned TSFM (0.013 in Table~\ref{tab:main-fc}).

\subsection{Other choices of the TSFM and the LLM}
\label{app:comp-swap}

\textbf{Question.} Section~\ref{sec:setup} instantiates \model{} with Chronos-2 and Qwen3.5-9B. This section replaces one component at a time and measures whether the results of Section~\ref{sec:results-main} depend on that choice.

\textbf{Protocol.} Two configurations are trained, each named by its two components as TSFM + LLM; in this naming \model{} is Chronos-2 + Qwen3.5-9B. TimesFM-3.0 + Qwen3.5-9B replaces the TSFM with TimesFM-3.0 \citep{timesfm}. Chronos-2 + Qwen3.5-4B replaces the LLM with Qwen3.5-4B, a smaller model of the same generation. Each configuration repeats the training of the corresponding \model{} checkpoint (Appendix~\ref{app:hyper}) with the same data, hyperparameters and GPU count, and is scored as in Section~\ref{sec:setup}. On forecasting, TimesFM-3.0 is also evaluated zero-shot and fine-tuned without the LLM, so that each coupled configuration is compared with its own TSFM. Chronos-2 + Qwen3.5-4B shares the Chronos-2 rows, which involve no LLM. TimesFM-3.0 outputs nine quantile levels and Chronos-2 outputs 21. The CRPS of Table~\ref{tab:main-fc} averages the quantile loss over the 21 levels, so the nine TimesFM-3.0 quantiles are linearly interpolated to these levels, and the four outermost levels are extrapolated.

\begin{table}[h]
\centering
\footnotesize
\caption{\model{} with another TSFM or another LLM. Each configuration is named by its two components, TSFM + LLM; Chronos-2 + Qwen3.5-9B is \model{}, and its numbers repeat Tables~\ref{tab:main-und} and~\ref{tab:main-fc}. \textbf{Top:} understanding, per-domain scores and their mean (columns as in Table~\ref{tab:main-und}). \textbf{Bottom:} forecasting, CRPS over the full horizon (columns as in Table~\ref{tab:main-fc}); a row that names a TSFM alone is that TSFM without the LLM, zero-shot or fine-tuned ($\dagger$).}
\label{tab:app-comp-swap}
\setlength{\tabcolsep}{3pt}
\resizebox{\linewidth}{!}{%
\begin{tabular}{lccccccccccccc}
\toprule
Understanding & Phys. & Tele. & Synth. & Urban. & IT & Neuro. & Earth. & Meteo. & Econ. & Bioac. & Manuf. & Ener. & \textbf{Avg.} \\
\midrule
Chronos-2 + Qwen3.5-9B & 65.0 & 97.0 & 97.4 & 96.2 & 84.1 & 98.5 & 73.8 & 97.0 & 80.6 & 78.5 & 64.6 & 78.1 & 84.2 \\
Chronos-2 + Qwen3.5-4B & 63.6 & 96.8 & 97.3 & 95.4 & 82.2 & 98.6 & 71.5 & 96.0 & 77.4 & 78.5 & 65.0 & 75.6 & 83.2 \\
TimesFM-3.0 + Qwen3.5-9B & 62.5 & 97.2 & 96.3 & 95.7 & 80.5 & 98.1 & 64.6 & 95.4 & 78.8 & 76.0 & 59.2 & 70.6 & 81.2 \\
\bottomrule
\end{tabular}}

\vspace{4pt}
\setlength{\tabcolsep}{5pt}
\begin{tabular}{lcccccc}
\toprule
Forecasting & Solar & Load & Traffic & Finance & Climate & \textbf{Avg.} \\
\midrule
Chronos-2 (zero-shot) & 0.511 & 0.066 & 0.187 & 0.032 & 0.065 & 0.172 \\
Chronos-2$^\dagger$ & 0.378 & 0.037 & 0.138 & 0.025 & 0.065 & 0.129 \\
Chronos-2 + Qwen3.5-9B & 0.357 & 0.032 & 0.113 & 0.022 & 0.056 & 0.116 \\
Chronos-2 + Qwen3.5-4B & 0.385 & 0.033 & 0.109 & 0.023 & 0.059 & 0.122 \\
\midrule
TimesFM-3.0 (zero-shot) & 0.479 & 0.058 & 0.215 & 0.030 & 0.086 & 0.174 \\
TimesFM-3.0$^\dagger$ & 0.376 & 0.040 & 0.150 & 0.029 & 0.087 & 0.136 \\
TimesFM-3.0 + Qwen3.5-9B & 0.361 & 0.036 & 0.146 & 0.027 & 0.084 & 0.131 \\
\bottomrule
\end{tabular}
\end{table}

\textbf{Understanding.} Both configurations score above every other row of Table~\ref{tab:main-und} on average, whose best is 74.9 (Table~\ref{tab:app-comp-swap}, top). Chronos-2 + Qwen3.5-4B averages 83.2, 1.0 point below \model{}. The per-domain differences are at most 3.2 points; Manufacturing and Neuroscience rise, and Bioacoustics is unchanged. TimesFM-3.0 + Qwen3.5-9B averages 81.2, 3.0 points below \model{}. Three domains account for 1.8 of these points: Earth Science ($-9.2$), Energy ($-7.5$) and Manufacturing ($-5.4$); on the other nine domains the difference is at most 3.5 points. Replacing the TSFM lowers the understanding average by 3.0 points, and replacing the LLM by 1.0 point.

\textbf{Forecasting.} The coupling lowers the CRPS of the fine-tuned TSFM under both TSFMs and both LLMs (Table~\ref{tab:app-comp-swap}, bottom). The coupled configuration has the lower CRPS on 14 of the 15 domain cells; the exception is Solar for Chronos-2 + Qwen3.5-4B. The gain is 0.013 for \model{}, 0.007 for Chronos-2 + Qwen3.5-4B and 0.005 for TimesFM-3.0 + Qwen3.5-9B. The two TSFMs are close without the LLM, at 0.172 and 0.174 zero-shot and at 0.129 and 0.136 fine-tuned. Replacing the LLM costs less than replacing the TSFM (0.006 against 0.015 CRPS). All three coupled configurations have a lower CRPS than the best time series model of Table~\ref{tab:main-fc} (0.170).

\end{document}